%% file: main.tex
\documentclass{article}

\usepackage[nonatbib,final]{neurips_2024}
\usepackage[utf8]{inputenc} 
\usepackage[T1]{fontenc}    
\usepackage{hyperref}       
\usepackage{url}            
\usepackage{booktabs}       
\usepackage{amsfonts}       
\usepackage{nicefrac}       
\usepackage{microtype}      
\usepackage{xcolor}         
\usepackage{subcaption}
\usepackage{graphicx}
\usepackage{algorithm2e}
\usepackage{multirow}

\input{sections/cmds}

\title{Impact of Label Noise on Learning Complex Features}

\author{%
  Rahul Vashisht\textsuperscript{1}\thanks{Equal contribution with random order.}\,\, ~~~
  P. Krishna Kumar \textsuperscript{1}\footnotemark[1]\,\, ~~~
  Harsha Vardhan Govind\textsuperscript{2}\thanks{Work done while at RBCDSAI, IIT Madras.} \\
  \textbf{Harish G. Ramaswamy}\textsuperscript{1}\\ \\
  \textsuperscript{1} Indian Institute of Technology Madras \textsuperscript{2}
  IIITDM Kancheepuram \\
    \texttt{\{rahul,pkrishna,hariguru\}@cse.iitm.ac.in},
  \texttt{cs21b1052@iiitdm.ac.in}\\
}

\begin{document}
\vspace{-0.1in}
\maketitle

\input{sections/abstract}

\input{sections/introduction}
\input{sections/methods}
\input{sections/experiments}
\input{sections/discussion}

\vspace{8mm}
\bibliographystyle{plain}
\bibliography{refs}

\appendix
\input{sections/appendix}

\end{document}

%% file: sections/cmds.tex
\newcommand\data{\mathcal{D}}

\newcommand\nostd[2]{$#1 ~{\scriptstyle \pm #2}$}

%% file: sections/abstract.tex
\vspace{-0.1in}
\begin{abstract}
\vspace{-0.1in}

Neural networks trained with stochastic gradient descent exhibit an inductive bias towards simpler decision boundaries, typically converging to a narrow family of functions, and often fail to capture more complex features.
This phenomenon raises concerns about the capacity of deep models to adequately learn and represent real-world datasets. 
%
Traditional approaches such as explicit regularization, data augmentation, architectural modifications, etc., 
have largely proven ineffective in encouraging the models to learn diverse features. 
%
%
%
In this work, 
we investigate the impact of pre-training models with noisy labels on the dynamics of SGD across various architectures and datasets. 
We show that pretraining promotes learning complex functions and diverse features in the presence of noise. 
Our experiments demonstrate that pre-training with noisy labels  encourages gradient descent to find alternate minima that do not solely depend upon simple features, rather learns more complex and broader set of features, without hurting performance. 


\end{abstract}

%% file: sections/introduction.tex
\vspace{-0.1in}
\section{Introduction}
\vspace{-0.1in}


Overparameterized models trained using stochastic gradient descent tend to focus on only a small fraction of the available features.
Such behavior reduces the diversity of features that contribute to the classification of data.
This phenomenon has been discussed in the context of \textit{simplicity bias} \cite{DBLP:conf/nips/ShahTR0N20,10.5555/3540261.3540358}, where neural models solely rely on simpler, easy-to-learn, features.  
It is also studied in relation to the intrinsic regularization properties of SGD that inherently favors lower-complexity models \cite{DBLP:conf/nips/KalimerisKNEYBZ19}. 
Previous works have shown that models trained using SGD learn linear functions first, and as training progresses they learn  functions of increasing complexity \cite{DBLP:conf/nips/KalimerisKNEYBZ19}.
%
This preference causes many potentially useful features to remain unlearned and underutilized, resulting in models that have inferior discriminative quality and rely on features that are coincidental rather than causal \cite{arjovsky2019invariant}. 
Such models when faced with distributional shifts or adversarially perturbed data, are unable to generalize to novel or altered environments.

%


%
%
%
Recent works have attributed this biased behavior to the intrinsic regularization properties of SGD that naturally favors low-complexity solutions \cite{DBLP:conf/nips/KalimerisKNEYBZ19}. 
Models that exhibit poor generalization for minority groups are shown to dedicate excess parameters to memorizing a small number of data points \cite{10.5555/3524938.3525711}.
Complex features are often overshadowed by the amplification and replication of simpler features \cite{addepalli2023feature}. 
In real-world datasets, 
simplicity bias in neural networks is often defined as the propensity to learn low-dimensional projections of inputs \cite{10.5555/3666122.3666475},
 where features such as shape and color (e.g., in MNIST or CIFAR)  define decision boundaries of varying complexity. 
Standard approaches like ensembling and adversarial training fail to effectively address the limitations imposed by this bias \cite{DBLP:conf/nips/ShahTR0N20}. 
%
%
There is an increasing interest in understanding the factors that contribute to SGD's bias toward simpler feature sets, \textit{shortcut feature learning}, as well as strategies to mitigate or exploit this tendency to improve model's performance. 
These include approaches that aim to reduce feature correlation \cite{addepalli2023feature,vasudeva2023mitigatingsimplicitybiasdeep,10.5555/3666122.3666475,Teney_2022_CVPR},
%
and use empirical risk minimization to learn more complex functions \cite{10.5555/3600270.3603061}.
More advanced methods that regularize the conditional mutual information of simpler models compel them to utilize a broader range of features, which has been shown to enhance out-of-distribution generalization \cite{vasudeva2023mitigatingsimplicitybiasdeep}.
%
SGD can also learn poorly generalizable functions when the models parameters are
initialized by training on random labels \cite{DBLP:conf/nips/LiuPA20}.
Similarly, the features selected across data points are shown to be strongly correlated with neural feature matrix (NFM) \cite{Radhakrishnan2022MechanismOF,Beaglehole2023MechanismOF}.
In the context of convolutional neural networks (CNNs), filter covariance is shown to closely mimic the average gradient outer product concerning input \cite{Beaglehole2023MechanismOF}. 
Despite these efforts, this bias remains a critical area of concern, necessitating further exploration to develop more robust and  OOD generalizable neural models. 

Pretraining with noisy labels is known to alter the optimization trajectory and often change the local minima to which SGD converges \cite{10.5555/3540261.3542363}. 
While adding noise is shown to help models generalize better, it has a tendency to overfit in overparameterized models \cite{frei2021provable}. 
The effects of adding noise is shown to be equivalent of employing a regularized loss function that depends on factors such as noise strength, batch size, etc.  \cite{10.5555/3540261.3542363,huh2024a}
Noisy labels have also been proposed for robust-loss-functions but they are often insufficient to learn accurate models \cite{wang2021learning}.
However, all these works do not explore the qualitative nature of diverse features that models learn. 

In this study, we investigate the learned decision-boundaries through the lens of feature diversity. 
We particularly focus on the role of parameter initialization in shaping the set of features utilized for classification. 
Our empirical analysis shows that: 
%
\begin{itemize}
	\item A simple pre-training phase by optimizing log-loss over perturbed \textit{noisy-labels} can be utilized to learn a more diverse family of functions by neural models. This is in contrast to the usual convergence of models to a similar family of decision functions.
	\item Neural models are capable of leveraging a broader array of features that are likely to exhibit greater generalization when faced with distribution shifts. 
\end{itemize}

%% file: sections/methods.tex
\vspace{-0.1in}
\section{Effect of Noisy Pre-training on Learned Decision Functions}
\vspace{-0.1in}
\label{sec:method}
We study a two stage training procedure: First, we pre-train an overparameterized model on corrupted labels (by randomly flipping the labels for a fraction of data points) called as \textbf{noisy pre-training}. 
This pre-trained model is then optimised to minimize the training loss, which naturally leads to a high training accuracy over the noisy-labels. 
Second, we utilize the pretrained model to once again train over the original unmodified labels. In subsequent sections, we compare this two-stage method with standard training without explicit regularization. 

\begin{figure}[t]
        \centering
     \begin{subfigure}{0.48\textwidth}
      \centering
      \includegraphics[width=\textwidth, height=1.9cm]{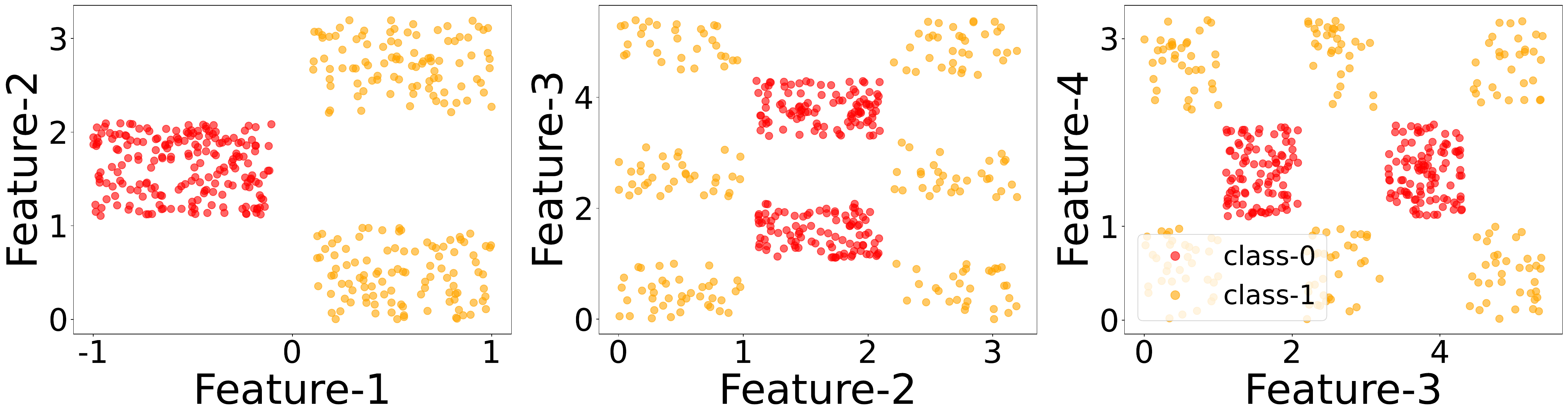}
        \label{fig1:subfig1}
    \end{subfigure}
      ~~~~
    \begin{subfigure}{0.47\textwidth}
      \centering
      \includegraphics[width=\textwidth]{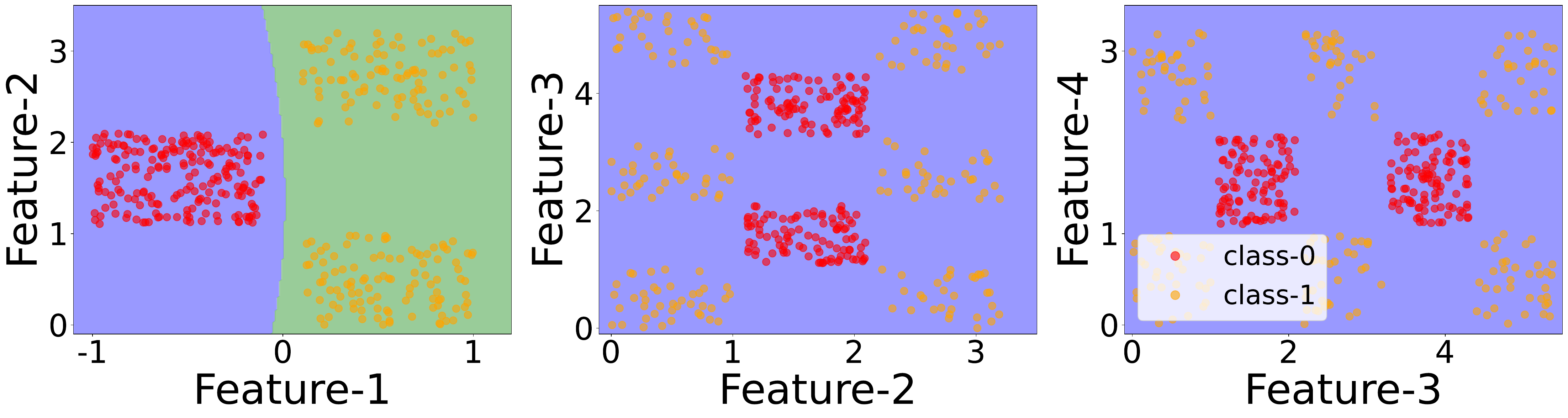}
        \label{fig4:subfig1}
    \end{subfigure}
        \label{fig1:my_label}
   \begin{subfigure}{0.48\textwidth}
      \centering
      \includegraphics[width=\textwidth]{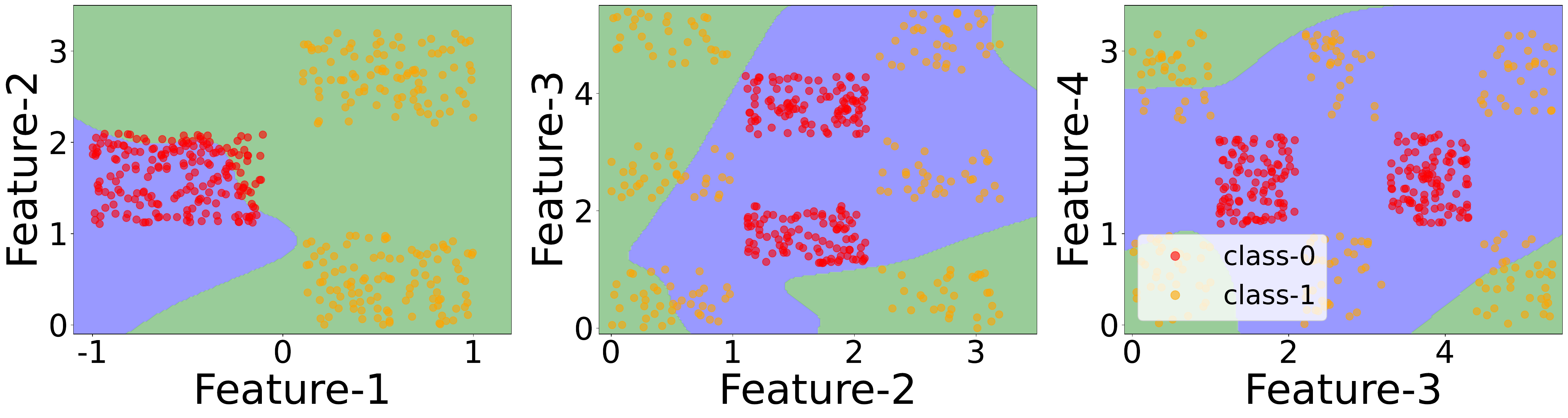}
        \label{fig4:subfig2}
    \end{subfigure}
    ~~~
    \begin{subfigure}{0.48\textwidth}
      \centering
      \includegraphics[width=\textwidth]{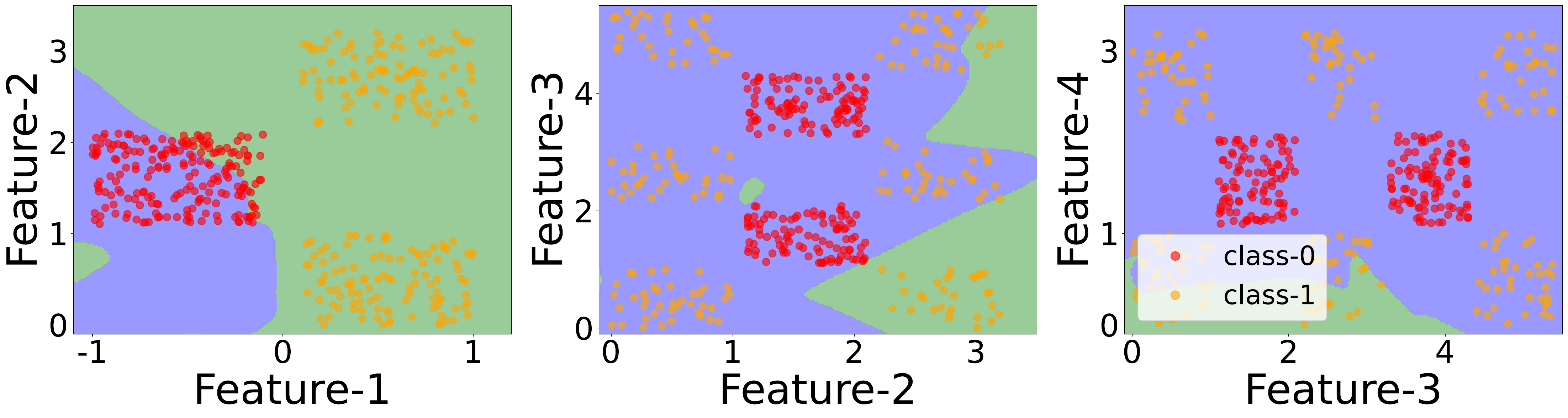}
        \label{fig4:subfig3}
    \end{subfigure}
	\vspace{-0.1in}    
    \caption{ \textit{(Top-left)} First 4 dimensions of Slab-data.
 \textit{(Top-right)} Decision boundary learned from regular training. \textit{(Botton row)}  Two different decision boundaries learned after noisy pre-training.  All models achieve $100$\% training and test accuracy.}
    \vspace{-0.2in}
        \label{fig:slab_data_all}
    \end{figure}

Consider a $d$-dimensional synthetic Multi-slab dataset $\data$, whose first coordinate is a linear block with instances for class-$0$ sampled from Uniform distribution in $[-1,0]$ and class-$1$ sampled from Uniform distribution in $[0,1]$. The remaining $d-1$ coordinates have samples from two classes distributed in $k$-well separated alternating regions \footnote{To eliminate any special association related to the ordering of features, we apply a random orthogonal projection to the dataset, and use the projected features instead of the original feature set.} \cite{DBLP:conf/nips/ShahTR0N20}. We consider a a $4$-dimensional multi-slab data with increasing complexity. Figure \ref{fig:slab_data_all} illustrates the $4$-dimensional multi-slab data.
Here, an overparameterized ReLU network can potentially use any feature to determine the class label. 
Note that, Feature-$1$ requires a simple linear decision function, whereas Feature-$2$ requires a $3-$piecewise linear function. 
Standard training leads to decision functions that solely depend on feature-$1$ and thus learn a linear function. 
While this is perfectly good for linearly separable data, this tendency persists even in cases with slight feature overlap in classes along Feature-1, diminishing the model's generalizability. 
This observation aligns with the well-documented phenomenon of \textit{shortcut learning} \cite{hermann2024foundationsshortcutlearning,murali2023distributionshiftspuriousfeatures,10.5555/3540261.3540358} and the behavior of neural networks as max-margin classifiers \cite{10.5555/3291125.3309632,NEURIPS2018_0e98aeeb}. 

In case of noisy pretraining, this sole dependence on Feature-1 is lost, and dependence on other feature increases as shown in the Figure \ref{fig:slab_data_all}. 
Randomized shuffle accuracy help understand the dependence on other features. Table \ref{tbl:multi-slab} shows the accuracy for randomly shuffling different features. 
Noisy pre-training increases the shuffle accuracy for feature-$1$ from $50.3\%$ to $64.2\%$, thus showing the increased dependence of the decision functions on other features. 
This is also evident from the reduced shuffle accuracy for feature-$2$ and feature-$3$ under noisy pre-training setting. 
The key observations are that the decision boundary is no longer solely reliant on the first dimension, but now incorporates information from multiple dimensions, 
and different noisy-datasets, $\data$ and $\data'$, result in distinct decision boundaries. This is unlike earlier instances where the model consistently converged to the same decision boundary.

\textbf{Significance of this observation:}
The benefits of moving beyond the simpler models typically learned by SGD-based optimizers are evident for improved \textit{out-of-distribution} generalization. 
Rather than basing classification decisions on a single feature, the model begins to leverage a broader range of features.
Additionally, this result provides evidence that there are multiple local minima with similar levels of accuracy across $\data$, and that SGD is capable of converging to these diverse solutions, rather than being confined to a single-simple decision boundary.

%% file: sections/experiments.tex
\section{Experiments}
\vspace{-0.1in}
In this section, we empirically demonstrate the impact of initializing parameters from noisy pretraining over Dominoes \cite{DBLP:conf/nips/ShahTR0N20} and Waterbirds data \cite{ Petryk2022OnGV}. 
See Appendix \ref{datadesc} for more datasets.

\textbf{Datasets:}~~
 \textit{MNIST-fMNIST} dataset from Dominoes consists of collated images, where $\{0,1\}$ MNIST digits are vertically stacked with \{\textit{shirt, frock}\} Fashion-MNIST apparel.
Here, MNIST block is the easier feature to learn, whereas fMNIST is complex.
WaterBirds dataset comprises of birds natively found on land/water, with different backgrounds of land or water. 
An image of water-bird with water in the background (similarly land-bird on land) is called \textit{in-group} partition,
whereas mismatch of waterbird-on-land or landbird-on-water is referred to as \textit{out-group} partition.
Details in Appendix \ref{datadesc}.

\textbf{Measuring feature dependence:}~~
To identify the feature(s) that the learned model is utilizing for  classification decision, we use:  
\textbf{(1).} \textit{Randomized shuffle accuracy}: to selectively shuffle one among MNIST (top) or fMNIST (bottom) part of the test images, and reevaluate the model. 
If its performance is drastically reduced by shuffling a feature, it implies the dependence of model's decision on the corresponding feature. 
\textbf{(2).} \textit{Visualizing Gram matrix}:  We plot diagonal entries of Gram matrix $W_1^TW_1$ to observe changes in the first layer's parameter $W_1$. 
The brighter parts of $W_1^TW_1$ denote greater dependence on those corresponding pixels in the input image.
Further, Appendix \ref{sec:appndx_eigen} shows eigenvector based visualization for Gram matrix.



\textbf{Varying correlation between features \& class labels}: 
We control the predictive powers of features by perturbing their correlation with true labels.
For example: In Dominoes dataset $\data$, we create partially-correlated data $\data'$ by collating images such that top MNIST-block is only $95\%$ correlated with classification labels (not $100\%$ accurately predictive). 
This implies that if predictions were solely on the basis of top-block, then the bayes-error will be $5\%$.
The rationale of adding adversarial correlation is to push the models to those regimes where no simple solution achieves perfect accuracy, thus getting closer to real world scenarios. 

\renewcommand{\arraystretch}{1.2}
\begin{table}[h]
\centering
\begin{tabular}{|l|lll|}
\hline
\multicolumn{1}{|c|}{\multirow{2}{*}{Parameter Initialization Regime}} & \multicolumn{3}{c|}{Feature Shuffle Accuracy}                                                                            \\ \cline{2-4} 
\multicolumn{1}{|c|}{}                       & \multicolumn{1}{c|}{Feature-1}           & \multicolumn{1}{c|}{Feature-2}          & \multicolumn{1}{c|}{Feature-3} \\ \hline
Standard data-agnostic (like Xavier-glorot) & \multicolumn{1}{l|}{\nostd{50.3}{1.78}} & \multicolumn{1}{l|}{\nostd{100}{0.0}}  & \nostd{100}{0.0}              \\ \hline
Pretraining model with $10\%$ noisy labels                                 & \multicolumn{1}{l|}{\nostd{64.2}{6.8}}  & \multicolumn{1}{l|}{\nostd{91.6}{4.3}} & \nostd{97.1}{2.2}             \\ \hline
\end{tabular}
 \vspace{0.1in}
\caption{\textbf{Multi-Slab Data:} Randomized feature shuffle accuracies, averaged over $10$ runs. All models are trained to achieve $100$\% training and test accuracy. 
 } \label{tbl:multi-slab}
\vspace{-0.2in}
\end{table}

\begin{table}[h]
\centering
\begin{tabular}{|c|ccc|ccc|}
\hline
\multirow{2}{*}{Data} &
  \multicolumn{3}{c|}{Standard Training} &
  \multicolumn{3}{c|}{Noisy Pre-training} \\ \cline{2-7} 
 &
  \multicolumn{1}{c|}{Train Acc.} &
  \multicolumn{1}{c|}{MNIST Rnd.} &
  F-MNIST Rnd. &
  \multicolumn{1}{c|}{Train Acc.} &
  \multicolumn{1}{c|}{MNIST Rnd.} &
  F-MNIST Rnd. \\ \hline
$\data$ &
  \multicolumn{1}{c|}{\nostd{99.9}{0.0}} &
  \multicolumn{1}{c|}{\nostd{52.5}{0.33}} &
  \nostd{98.3}{0.05} &
  \multicolumn{1}{c|}{\nostd{99.7}{0.07}} &
  \multicolumn{1}{c|}{\nostd{53.6}{1.56}} &
  \nostd{88.6}{0.76} \\ \hline
$\data'$ &
  \multicolumn{1}{c|}{\nostd{98.5}{0.07}} &
  \multicolumn{1}{c|}{\nostd{93.1}{0.33}} &
  \nostd{56.5}{0.42} &
  \multicolumn{1}{c|}{\nostd{99.9}{0.06}} &
  \multicolumn{1}{c|}{\nostd{81.2}{1.02}} &
  \nostd{57.2}{1.50} \\ \hline
\end{tabular}

 \vspace{0.1in}
    \caption{\textbf{Dominoes MNIST-FMNIST} (Rnd. short for Randomized) : 
    Randomized shuffle accuracies, averaged over 3 runs, with 100\% ($\mathcal{D}$) and 95\% ($\mathcal{D}'$) correlation of MNIST with the true labels. 
    Each for 3 and 4 layer fully connected networks. Noisy pre-training uses $10\%$ corrupt labels.  }  \label{tbl:dominoes}
    	\vspace{-0.2in}
\end{table}

\textit{Noisy Pre-training Helps Learn Complex Features}: ~~~
We observe that neural models usually learn simple decision functions that are restricted to \textit{easy-to-learn} features. 
Table\ref{tbl:multi-slab} shows that randomizing feature-1 reduces the accuracy to $50\%$, proving absolute reliance on this feature only. Similarly, Table\ref{tbl:dominoes} shows randomizing top MNIST block reduces model accuracy to $52.5\%$ and Table\ref{tbl:waterbird} shows better performance for in-group images. 
Conversely, the model does not deteriorate performance upon randomizing other complex features in all datasets.
However,  when we initialize parameters from noisy pretrained models, we see that not only over reliance of easy-features is reduced, the model's ignorance to other features is reduced. 
Table\ref{tbl:multi-slab} and \ref{tbl:dominoes} evince increase in randomized shuffling accuracy and Table\ref{tbl:waterbird} shows better out-group performance.

\textit{Noisy Pre-training Helps Learn Diverse Functions}: ~~~
With different label noise, we observe that the decision boundaries learned are varying across each feature. 
Figure\ref{fig:slab_data_all} bottom-row shows two different decision boundaries for varying noise during pretraining. 
Diverse models are helpful because they tend to focus on different aspects of of high dimensional data. 
In practical scenarios, we can make much more robust predictions by bagging these multiple decision boundaries to give aggregated class prediction.

\textit{Impact of Label Smoothing:}~~~ Please refer Appendix \ref{sec:appdx-smoothing}.

With noisy-pretraining and label smoothing, the models learn to focus on multiple features in Multi-slab and Dominoes data, and learn to classify images based on birds, rather than their backgrounds in case of WaterBirds. Figure \ref{fig:dominos} and \ref{fig:waterbirds} shows increase in the brightness of pixels in the Gram-matrix that confirms learning of diverse/complex features.  \\
\textit{Why are diverse models better?} ~
Model diversity is evident in slab-data from Figure\ref{fig15:show-all_random_init}.
It is interesting to note that each noisy pretraining leads to a different minina, and different family of function. 
The predictions from diverse models, focusing on diverse features, can be ensembled to have more robust predictions.
Similar trends are also evident from high standard-deviation values in accuracy of models with noisy pretraining in Table\ref{tbl:dominoes} and \ref{tbl:waterbird}.
The theoretical validation of such ensemble's superiority is for further exploration.

\begin{table}[h]
\centering
\begin{tabular}{|c|ccc|ccc|}
\hline
\multirow{2}{*}{Data} &
  \multicolumn{3}{c|}{Standard Training} &
  \multicolumn{3}{c|}{Noisy Pre-training} \\ \cline{2-7} 
 &
  \multicolumn{1}{c|}{Train Acc.} &
  \multicolumn{1}{c|}{In-group} &
  Out-group &
  \multicolumn{1}{c|}{Train Acc.} &
  \multicolumn{1}{c|}{In-group} &
  Out-group \\ \hline
$\data$ &
  \multicolumn{1}{c|}{\nostd{100}{0.0}} &
  \multicolumn{1}{c|}{\nostd{85.2}{0.43}} &
  \nostd{38.5}{0.88} &
  \multicolumn{1}{c|}{\nostd{99.5}{0.34}} &
  \multicolumn{1}{c|}{\nostd{78.1}{1.02}} &
  \nostd{44.1}{1.60} \\ \hline
$\data'$ &
  \multicolumn{1}{c|}{\nostd{100}{0.0}} &
  \multicolumn{1}{c|}{\nostd{84.1}{0.48}} &
  \nostd{44.4}{0.67} &
  \multicolumn{1}{c|}{\nostd{99.5}{0.73}} &
  \multicolumn{1}{c|}{\nostd{77.8}{1.15}} &
  \nostd{46.9}{0.92} \\ \hline
\end{tabular}
 \vspace{0.1in}
\caption{\textbf{WaterBirds Dataset}: Test In-group and Out-group accuracies averaged over 10 runs on waterbirds dataset, with 100\% ($\mathcal{D}$) and 95\% ($\mathcal{D}'$) correlation of background with the true labels. For noisy pre-training, we corrupt $10\%$ of labels. }  \label{tbl:waterbird}
\end{table}

%% file: sections/discussion.tex
\section{Discussion and Conclusion}
\vspace{-0.1in}
Overparameterized neural networks tend to learn decision functions based on a limited and simpler set of features if they exist. 
Our experiments demonstrate that initializing the model parameters by first training over noisy-labels can enable the model to overcome this constraint, and allow for the learning of more diverse range of features.
One possible explanation for this phenomenon is that the neural model reaches a minima that is characterized by a complex decision boundary to fit the noisy labels. 
The model, however, remains trapped in this minima and is unable to revert to a simpler decision boundary despite the intrinsic regularization properties of SGD. 
This observation challenges the commonly accepted opinion that SGD provides effective implicit regularization.
%

Our experiments motivate further investigation into the nature of the role of initialization in guiding SGD optimizers. 
Note that all the models, with or without noisy label pretraining, have similar (high) accuracy on unseen data.
Since each local minima that is reached during the noisy pre-training leads to a distinct family of functions, it hints at  interesting future works about characterizing loss-landscape of ReLU networks. 
This is in contrast to the standard data-agnostic initializations, such as Xavier-Glorot, that converge to similar decision boundaries.
%
%
Our work connects feature subset selection with the learning of diverse families of functions.
Future works will explore any theoretical relationships between these concepts and to quantify their interdependency.

Our observations challenge the prevalent view in the literature that neural networks are extremely biased towards learning only simple features if they exist. 
We demonstrate that models can be guided to learn more complex features through pre-training on noisy labels. 
As a corollary, we show that the phenomenon of \textit{shortcut learning} is not as extreme as previously thought and can be mitigated to some extent. 
Our experiments further reveal that neural networks are capable of learning multiple, higher-complexity features, and diverse families of functions.


\begin{figure}[h]
        \centering
     \begin{subfigure}{0.88\textwidth}
      \centering
      \includegraphics[width=\textwidth, height=4cm]{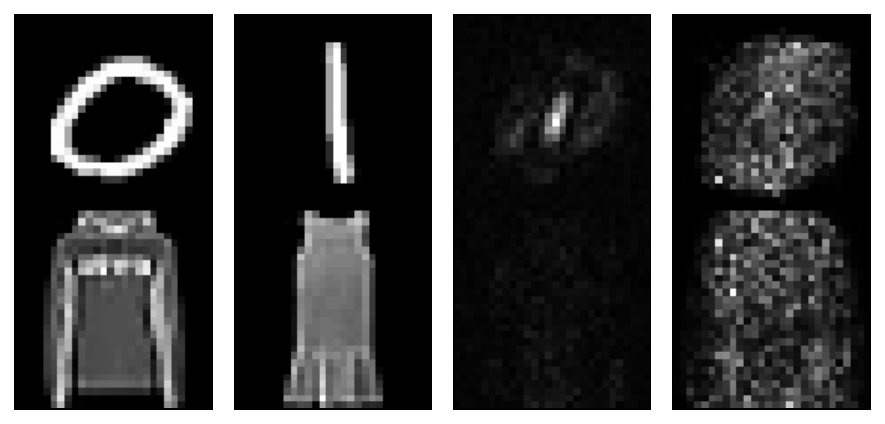}
        \label{fig:fmnist}
    \end{subfigure} 
    \caption{ MNIST-FMNIST dominoes data samples with visualizing first layer neural feature matrix $W_1^{\top}W_1$, for \textbf{(Left)} Standard training, and \textbf{(Right)} Random pre-training respectively. 
    }
    \vspace{-0.2in}
        \label{fig:dominos}
    \end{figure}


\begin{figure}[h]
    \centering
    \includegraphics[width=0.24\linewidth]{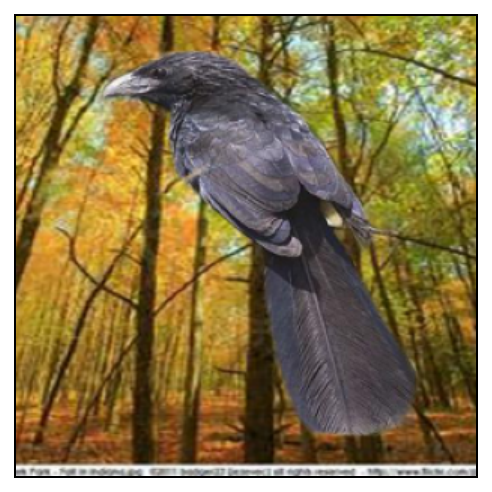}
    \includegraphics[width=0.24\linewidth]{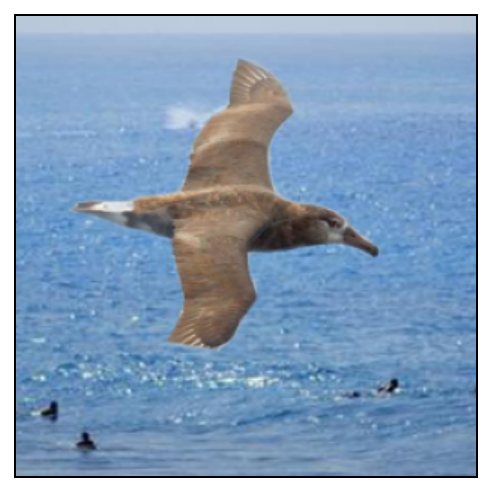}
    \includegraphics[width=0.24\linewidth]{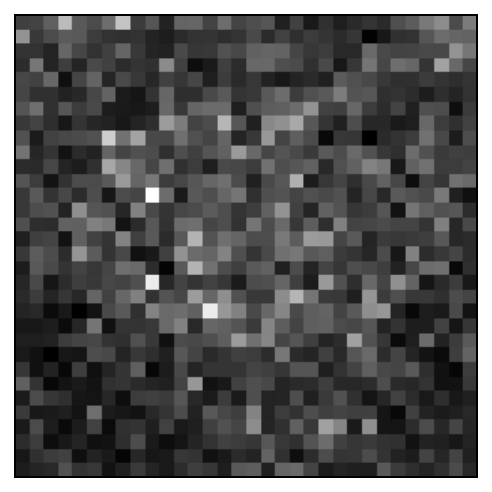}
    \includegraphics[width=0.24\linewidth]{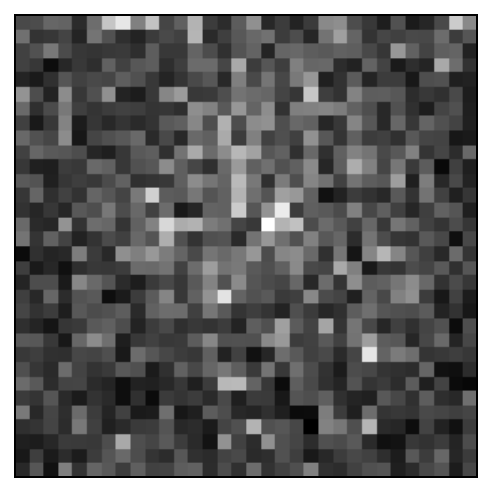}
    \caption{Samples from Waterbirds dataset. We visualize the diagonal of first layer neural feature matrix $W_1^{\top}W_1$ for \textbf{(Left)} standard training and \textbf{(Right)} random pretraining.}
    \vspace{-0.2in}
    \label{fig:waterbirds}
\end{figure}

%% file: sections/appendix.tex
\section{Appendix}
\label{sec:appdx}

\subsection{Dataset Description:}

\label{datadesc}
\begin{figure}[!h]
        \centering
     \begin{subfigure}[b]{0.32\textwidth}
      \centering
      \includegraphics[width=\textwidth]{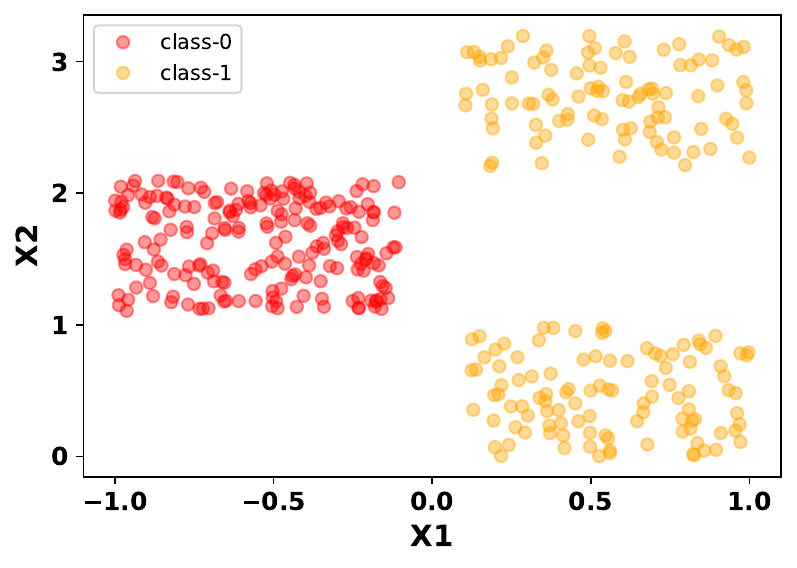}
        \label{fig2:subfig1}
    \end{subfigure}
    \begin{subfigure}[b]{0.32\textwidth}
      \centering
      \includegraphics[width=\textwidth]{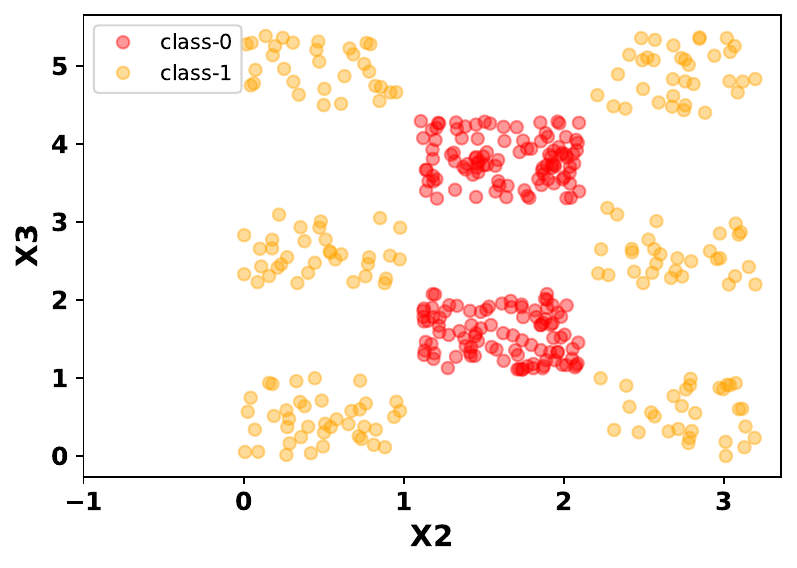}
        \label{fig2:subfig2}
    \end{subfigure}
         \begin{subfigure}[b]{0.32\textwidth}
      \centering
      \includegraphics[width=\textwidth]{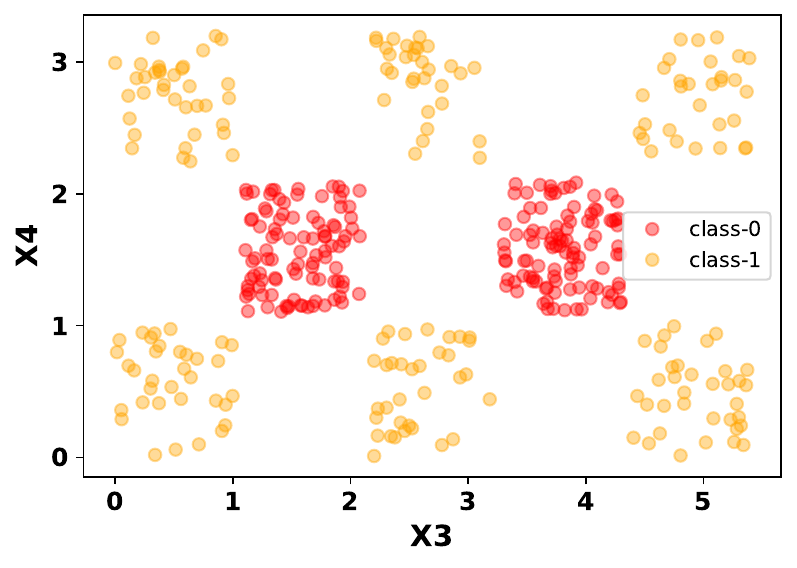}
        \label{fig2:subfig3}
    \end{subfigure}
    \caption{Noise augmented multi-slab dataset}
        \label{fig4dclean}
    \end{figure}

\begin{figure}[!h]
        \centering
     \begin{subfigure}[b]{0.32\textwidth}
      \centering
      \includegraphics[width=\textwidth]{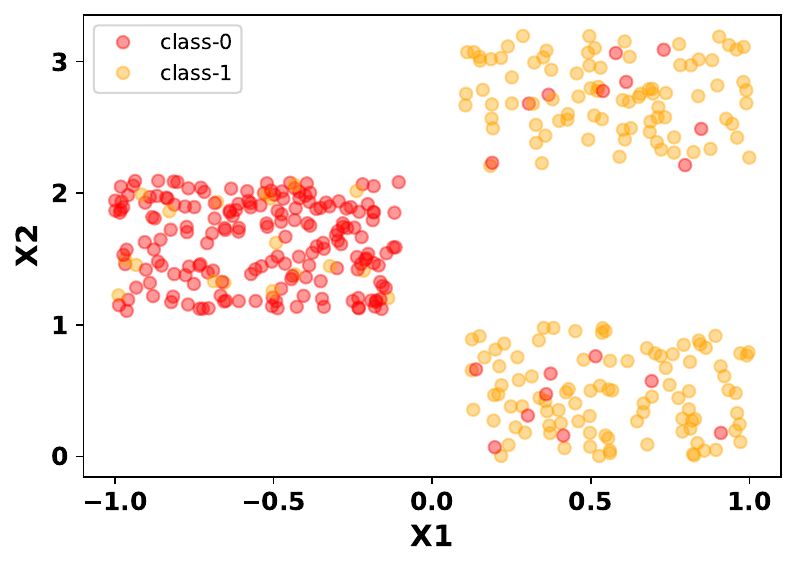}
        \label{fig2:subfig1}
    \end{subfigure}
    \begin{subfigure}[b]{0.32\textwidth}
      \centering
      \includegraphics[width=\textwidth]{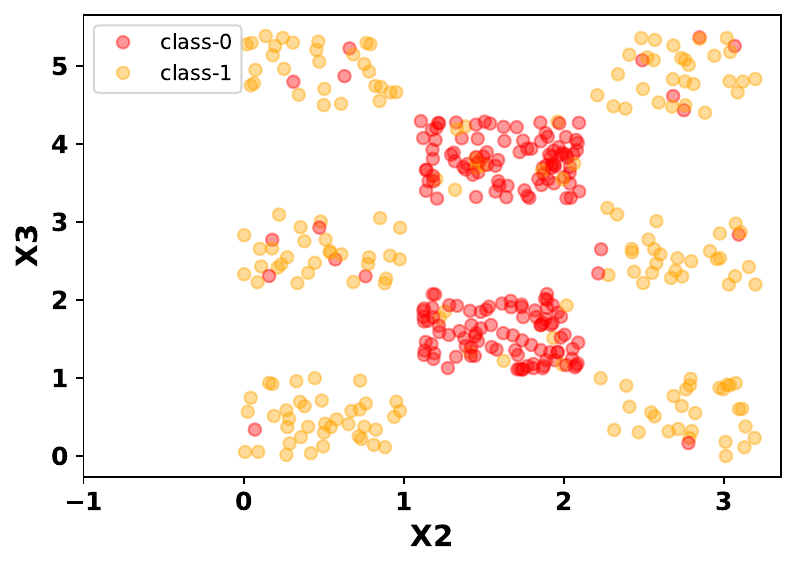}
        \label{fig2:subfig2}
    \end{subfigure}
         \begin{subfigure}[b]{0.32\textwidth}
      \centering
      \includegraphics[width=\textwidth]{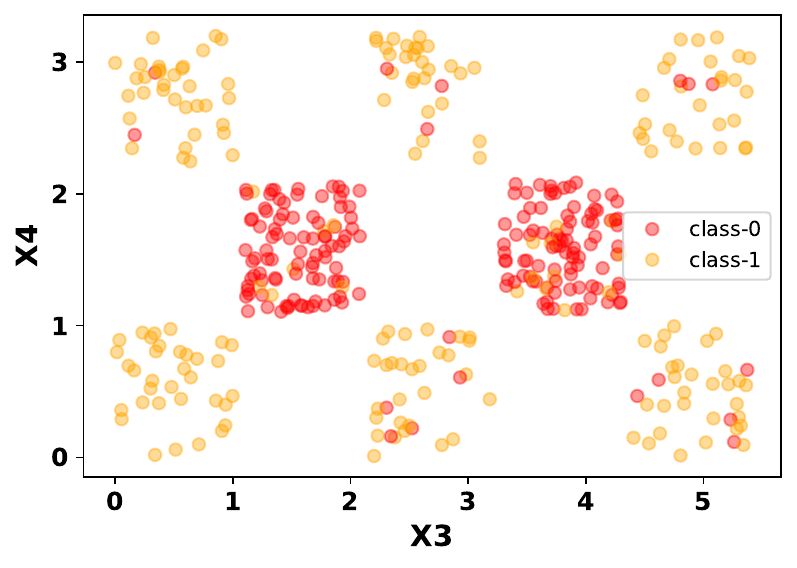}
        \label{fig2:subfig3}
    \end{subfigure}
    \caption{Noise augmented multi-slab dataset}
        \label{fig4dnoisy}
    \end{figure}

\textbf{Slab Data:} Figures \ref{fig4dclean} and \ref{fig4dnoisy} shows 4-dimensional dataset and noisy 4-dimensional multi-slab dataset.

\begin{figure}[!ht]
    \centering
    \includegraphics[width=\linewidth]{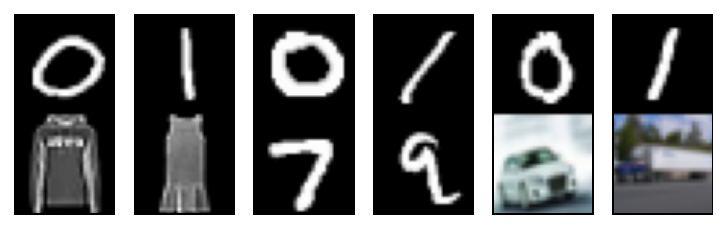}
    \caption{Samples from Dominoes Dataset, MNIST-FMNIST, MNIST-MNIST, and MNIST-CIFAR }
    \label{fig:d}
\end{figure}

\textbf{Dominoes Data:} We consider Dominoes binary classification datasets consisting of 3 independent datasets, where the top half of the image contains MNIST digits \cite{LeCun2005TheMD} from classes {$0$, $1$}, and the bottom half contains MNIST images from classes {$7$, $9$} (MNIST-MNIST), Fashion-MNIST \cite{DBLP:journals/corr/abs-1708-07747} images from classes {coat, dress} (MNIST-Fashion) or CIFAR-10 \cite{cifar10} images from classes {car, truck} (MNIST-CIFAR). In all Dominoes datasets, the top half of the image (MNIST 0-1 images) presents a easy to learn feature; the bottom half of the image presents a harder-to-learn feature. The images are made into gray-scale and scaled to appropriate size so that they can be collated top-bottom style.  Figure \ref{fig:d} shows the examples of dominoes dataset. In case of 100\% correlation Note that each block, top and bottom, is fully predictive of the class label in case of 100\% correlation, while only harder to learn feature is fully predictive of class label in case of 95\% correlation. 

\begin{figure}[!ht]
    \centering
    \includegraphics[width=0.24\linewidth]{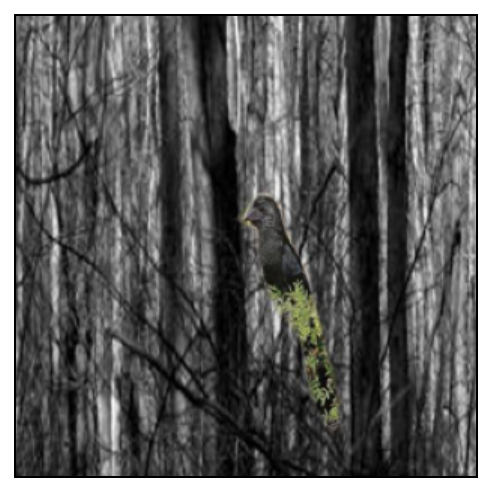}
    \includegraphics[width=0.24\linewidth]{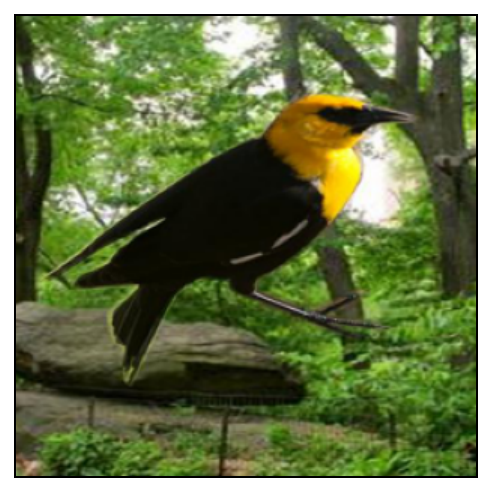}
    \includegraphics[width=0.24\linewidth]{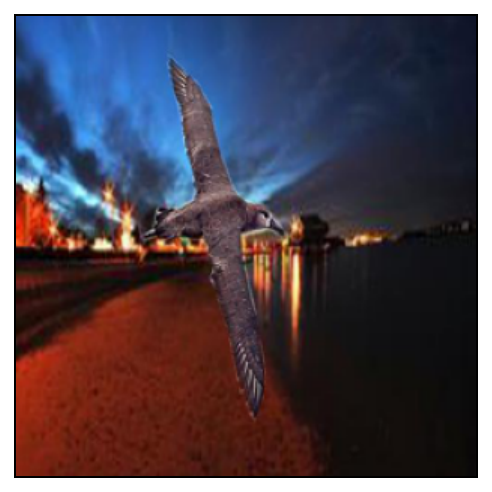}
    \includegraphics[width=0.24\linewidth]{figs/Waterbirds/waterbirds-sample-3.pdf}
    \caption{Samples from Waterbirds dataset }
    \label{fig:w}
\end{figure}

\textbf{Waterbirds Dataset:} The waterbirds dataset is constructed by cropping birds from Caltech-UCSD Birds-200-2011 (CUB) dataset \cite{Wah2011TheCB} and taking backgrounds from the Places dataset  \cite{7968387} and placing the birds from CUB dataset to backgrounds from Places dataset. Figure \ref{fig:w} shows sample instances from waterbirds dataset. Table \ref{tablewb1} describes the proportions of different groups in the waterbirds dataset. The waterbirds on land and landbirds on water are known as out-group, while landbird on land and waterbirds on water are known as the in-group.

\begin{table}[!ht]
    \centering
    \begin{tabular}{|c|c|c|c|c|}
    \hline
       Data Split  & Landbirds, Land & Landbirds, Water  & Waterbirds, land & Waterbirds, Water   
       \\
        \hline
               Train $\data$  &  3694 (77.04\%) & 0  & 0  &  1101 (22.96\%) \\ 
         Train $\data'$ &  3498 (72\%) & 184 (3.8\%) & 56 (1.2\%) &  1057 (22\%) \\ 
         Validation & 467 (38.95\%)& 466 (39.95\%) & 133 (11.1\%) & 133(11.1\%) \\
         Test & 2255 (38.92\%) & 2255(38.92\%) & 642 (11.1\%) & 642 (11.1\%)\\
         \hline
    \end{tabular}
    \vspace{0.1in}
    \caption{WaterBirds data description:  $\data$ and $\data'$ with $100\%$ and $95\%$  correlation respectively \cite{Petryk2022OnGV}}
    \label{tablewb1}
\end{table}

%

\subsection{Network Architectures}

\textbf{Slab Dataset:} We use 2-layer multi-layer perceptron network with ReLU activation, the first layer contains 100 hidden unit and second layer contains 200 hidden units. We use SGD with $0.1$ learning rate as optimizer.

\textbf{Dominoes Dataset:} For all dominoes dataset, ``MNIST-FMNIST'', ``MNIST-MNIST'', and ``MNIST-CIFAR'', datasets, we use 4-layer fully connected networks with 10 hidden units and ReLU activation. We use SGD optimizer with learning rate $0.01$. 

\textbf{WaterBirds Dataset:} We use fully connected network, with $4$ layers and ReLU activation having 20 hidden units. We use SGD optimizer with learning rate $0.05$ .

%
%
\subsection{Additional Results:}

\begin{table}[h]
\begin{tabular}{|c|ccc|ccc|}
\hline
\multirow{2}{*}{\begin{tabular}[c]{@{}c@{}}WaterBird\\ Dataset\end{tabular}} &
  \multicolumn{3}{c|}{Standard Training} &
  \multicolumn{3}{c|}{Noisy Pre-training} \\ \cline{2-7} 
 &
  \multicolumn{1}{c|}{Train Acc.} &
  \multicolumn{1}{c|}{In-group} &
  Out-group &
  \multicolumn{1}{c|}{Train Acc.} &
  \multicolumn{1}{c|}{In-group} &
  Out-group \\ \hline
$\data$ &
  \multicolumn{1}{c|}{\nostd{100}{0.0}} &
  \multicolumn{1}{c|}{\nostd{84.48}{0.59}} &
  \nostd{43.64}{3.56} &
  \multicolumn{1}{c|}{\nostd{99.89}{0.07}} &
  \multicolumn{1}{c|}{\nostd{77.77}{0.88}} &
  \nostd{49.89}{2.17} \\ \hline
$\data'$ &
  \multicolumn{1}{c|}{\nostd{100}{0.0}} &
  \multicolumn{1}{c|}{\nostd{83.48}{0.53}} &
  \nostd{49.44}{3.80} &
  \multicolumn{1}{c|}{\nostd{99.68}{0.08}} &
  \multicolumn{1}{c|}{\nostd{78.41}{0.56}} &
  \nostd{51.46}{1.48} \\ \hline
\end{tabular}
 \vspace{0.1in}
\caption{\textbf{WaterBirds Dataset}: Test In-group and Out-group accuracies averaged over 10 runs on waterbirds dataset, with 100\% ($\mathcal{D}$) and 95\% ($\mathcal{D}'$) correlation of background with the true labels. For noisy pre-training, we corrupt $10\%$ of labels. 
Label smoothing (LS) value is 0.1}
\end{table}

\begin{table}[h]
\begin{tabular}{|c|c|ccc|ccc|}
\hline
\multirow{2}{*}{\begin{tabular}[c]{@{}c@{}}MNIST \\ -MNIST\end{tabular}} &
  \multirow{2}{*}{\begin{tabular}[c]{@{}c@{}}Hidden \\ Dim\end{tabular}} &
  \multicolumn{3}{c|}{Standard Training} &
  \multicolumn{3}{c|}{Noisy Pre-training} \\ \cline{3-8} 
 &
   &
  \multicolumn{1}{c|}{Train Acc.} &
  \multicolumn{1}{c|}{Top Shuffle} &
  Bottom Shuffle &
  \multicolumn{1}{c|}{Train Acc.} &
  \multicolumn{1}{c|}{Top Shuffle} &
  Bottom Shuffle \\ \hline
\multirow{2}{*}{\begin{tabular}[c]{@{}c@{}}$\data$\\ \end{tabular}} &
  10 &
  \multicolumn{1}{c|}{100} &
  \multicolumn{1}{c|}{50.89} &
  99.75 &
  \multicolumn{1}{c|}{99.82} &
  \multicolumn{1}{c|}{53.20} &
  98.31 \\ \cline{2-8} 
 &
  50 &
  \multicolumn{1}{c|}{100} &
  \multicolumn{1}{c|}{50.15} &
  99.71 &
  \multicolumn{1}{c|}{100} &
  \multicolumn{1}{c|}{49.94} &
  99.50 \\ \hline
\multirow{2}{*}{\begin{tabular}[c]{@{}c@{}}$\data'$ \end{tabular}} &
  10 &
  \multicolumn{1}{c|}{98.81} &
  \multicolumn{1}{c|}{88.98} &
  59.81 &
  \multicolumn{1}{c|}{99.82} &
  \multicolumn{1}{c|}{90.81} &
  56.99 \\ \cline{2-8} 
 &
  50 &
  \multicolumn{1}{c|}{99.20} &
  \multicolumn{1}{c|}{90.74} &
  58.19 &
  \multicolumn{1}{c|}{99.92} &
  \multicolumn{1}{c|}{92.63} &
  57.76 \\ \hline
\end{tabular}
\vspace{0.1in}
\caption{\textbf{Dominoes MNIST-MNIST}: 
    Randomized shuffle accuracies, with 100\% ($\mathcal{D}$) and 95\% ($\mathcal{D}'$) correlation of MNIST with the true labels. \textbf{Top}:MNIST, \textbf{Bottom}:MNIST. 
    Each for 3 and 4 layer fully connected networks. Noisy pre-training uses $10\%$ corrupt labels.}
\end{table}

\begin{table}[h]
\begin{tabular}{|c|ccc|ccc|}
\hline
\multirow{2}{*}{\begin{tabular}[c]{@{}c@{}}MNIST \\ -CIFAR\end{tabular}} &
  \multicolumn{3}{c|}{Standard Training} &
  \multicolumn{3}{c|}{Noisy Pre-training} \\ \cline{2-7} 
 &
  \multicolumn{1}{c|}{Train Acc.} &
  \multicolumn{1}{c|}{MNIST Shuffle} &
  CIFAR Shuffle &
  \multicolumn{1}{c|}{Train Acc.} &
  \multicolumn{1}{c|}{MNIST Shuffle} &
  CIFAR Shuffle \\ \hline
$\data$ & \multicolumn{1}{c|}{100} & \multicolumn{1}{c|}{50.88} & 99.79 & \multicolumn{1}{c|}{99.66} & \multicolumn{1}{c|}{51.44} & 99.58 \\ \hline
$\data'$  & \multicolumn{1}{c|}{100} & \multicolumn{1}{c|}{54.62} & 87.14 & \multicolumn{1}{c|}{96.22} & \multicolumn{1}{c|}{53.48} & 92.30  \\ \hline
\end{tabular}
\vspace{0.1in}
\caption{\textbf{Dominoes MNIST-CIFAR}: 
    Randomized shuffle accuracies, with 100\% ($\mathcal{D}$) and 95\% ($\mathcal{D}'$) correlation of MNIST with the true labels. \textbf{Top}:MNIST, \textbf{Bottom}:CIFAR. Noisy pre-training uses $10\%$ corrupt labels.}
\end{table}

\begin{figure}[h]
\centering
    \begin{subfigure}{0.78\textwidth}
      \centering
      \includegraphics[width=\textwidth, height=5cm]{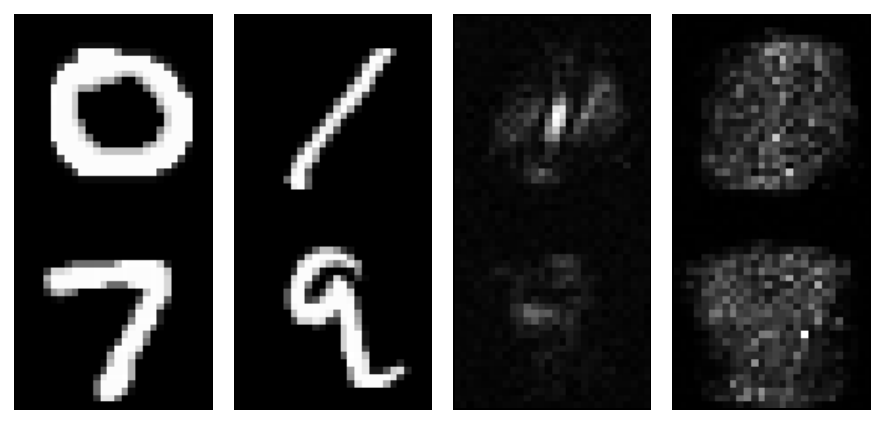}
        \label{fig:mnist}
    \end{subfigure} 
    ~~~~
        \begin{subfigure}{0.78\textwidth}
      \centering
      \includegraphics[width=\textwidth, height=5cm]{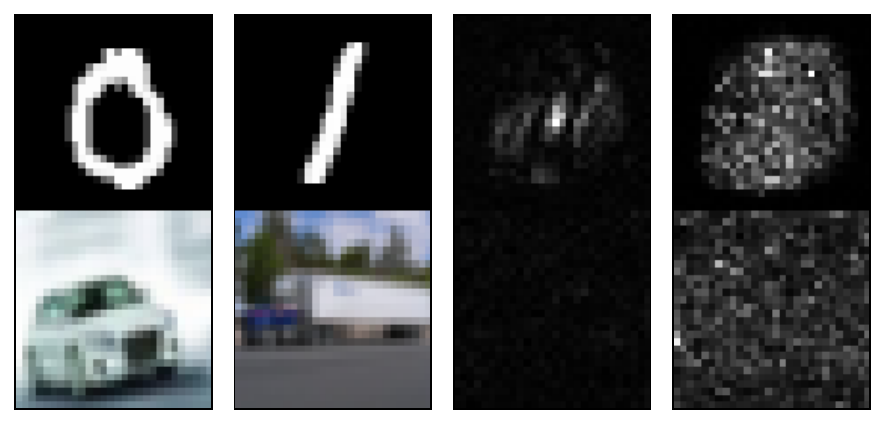}
        \label{fig:cifarmnist}
    \end{subfigure} 
    \caption{Visualizing the diagonal of first layer neural feature matrix $W_1^{\top}W_1$, where $W_1$ is parameters of first layer network of the learned network, \textbf{(Top)} MNIST-MNIST,  and \textbf{(Bottom)} MNIST-CIFAR Dominoes data samples with standard training and noisy pre-training respectively. }
    \label{fig:enter-label}
\end{figure}

\clearpage
\subsubsection{4D Slab Data Decision Boundary plots for Different Random Seeds}

\begin{figure}[h]
        \centering
        \begin{subfigure}{0.7\textwidth}
      \centering
      \includegraphics[width=\textwidth]{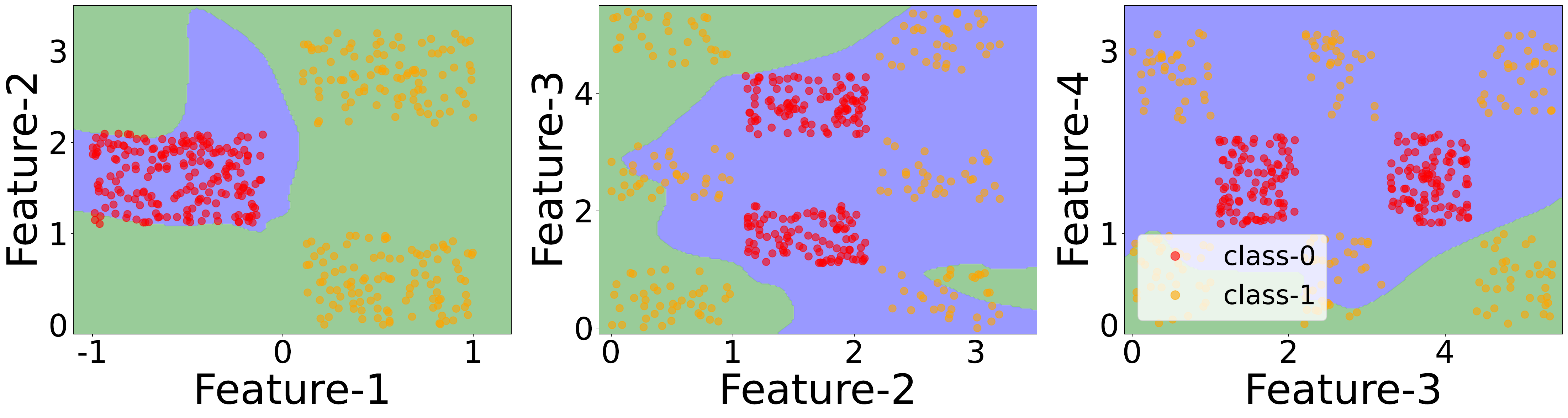}
    \end{subfigure}

   \begin{subfigure}{0.7\textwidth}
      \centering
      \includegraphics[width=\textwidth]{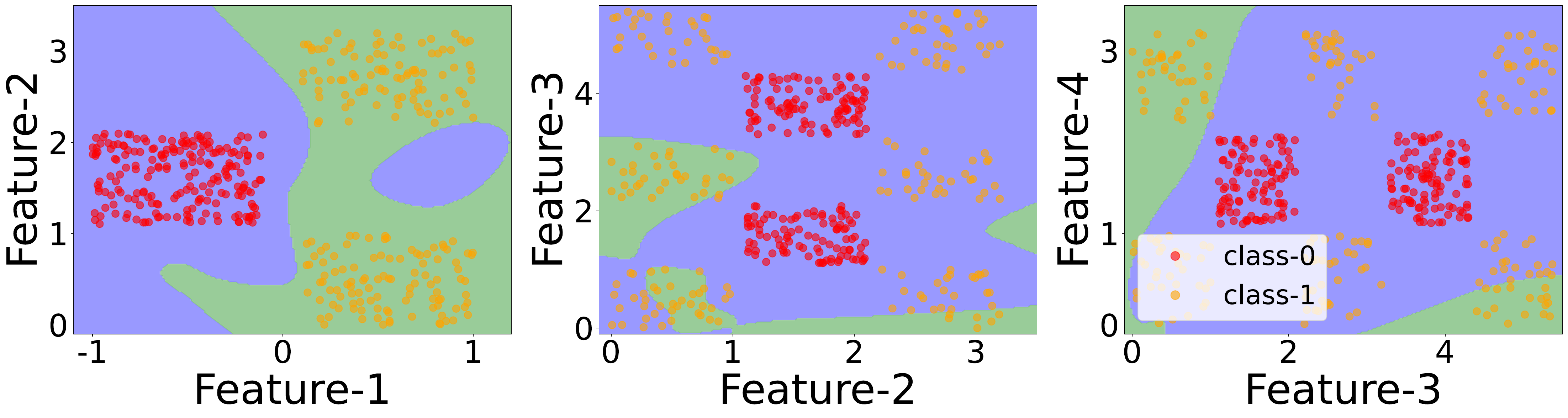}
    \end{subfigure}

 \begin{subfigure}{0.7\textwidth}
      \centering
      \includegraphics[width=\textwidth]{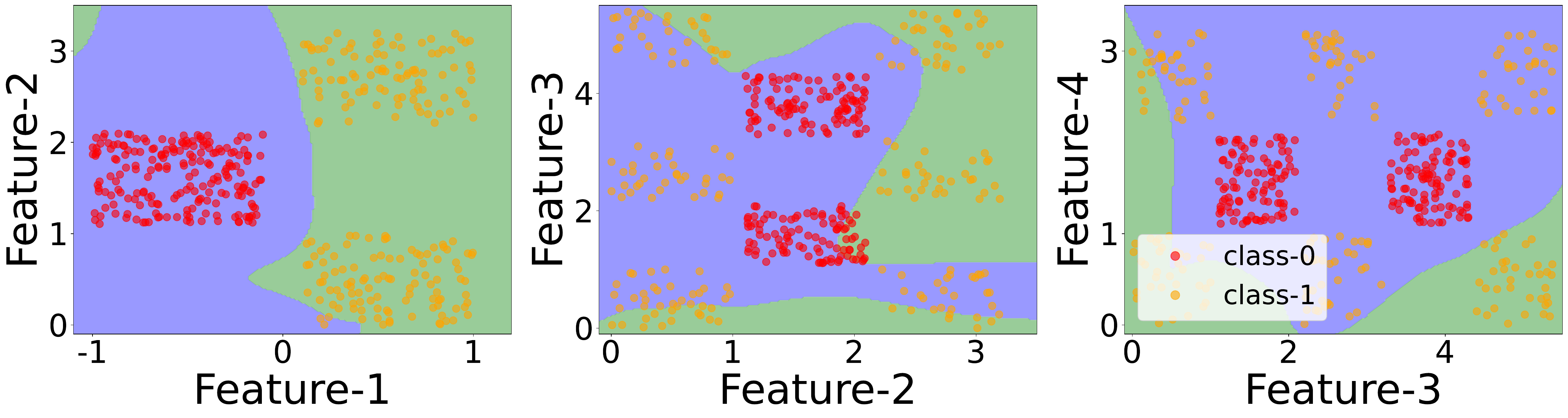}
    \end{subfigure}

      \begin{subfigure}{0.7\textwidth}
      \centering
      \includegraphics[width=\textwidth]{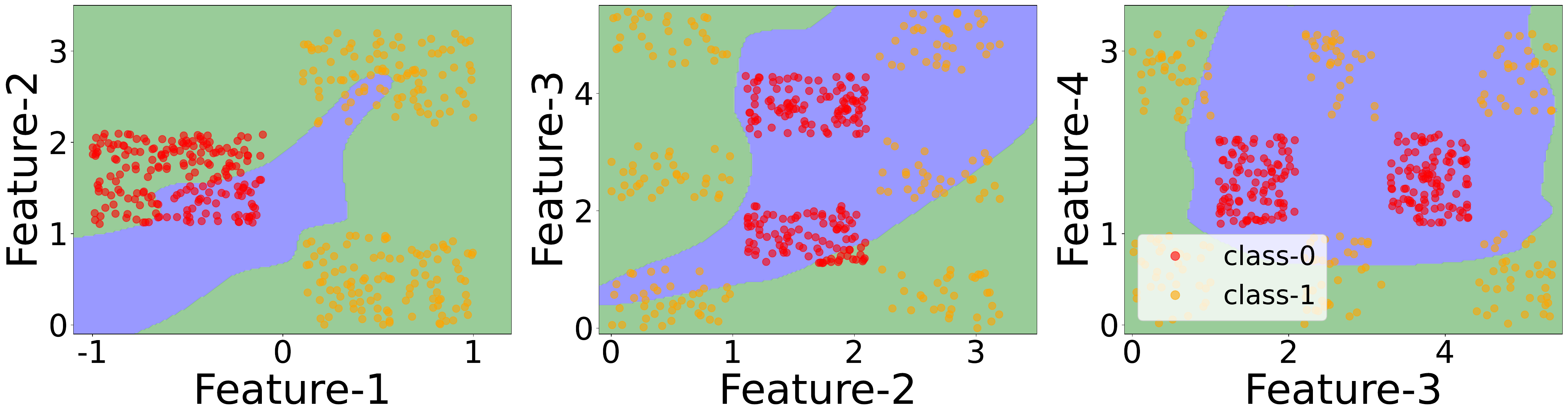}
    \end{subfigure}

    \begin{subfigure}{0.7\textwidth}
      \centering
      \includegraphics[width=\textwidth]{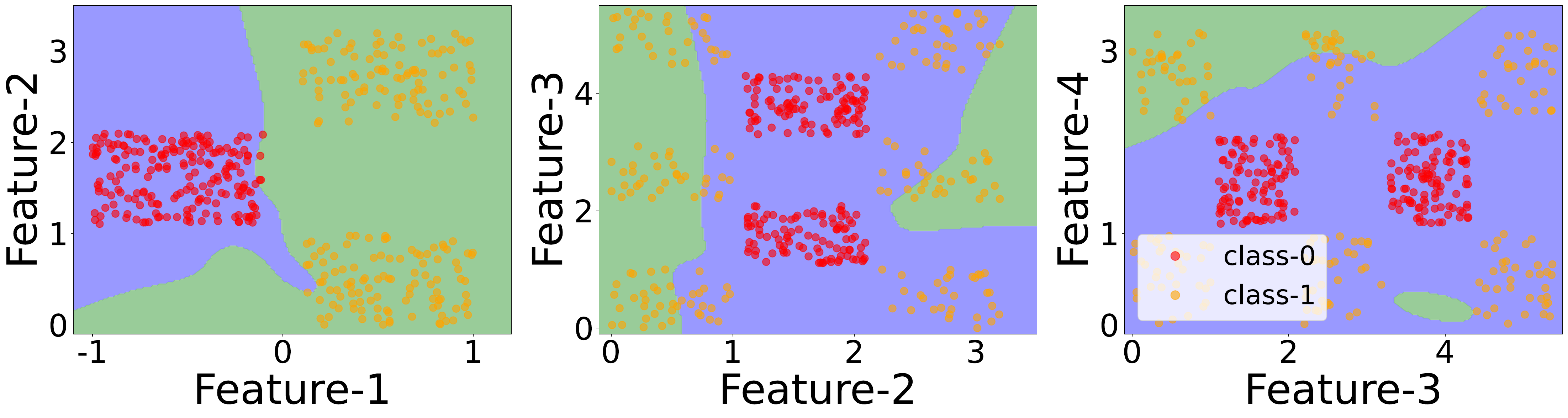}
    \end{subfigure}

    \begin{subfigure}{0.7\textwidth}
      \centering
      \includegraphics[width=\textwidth]{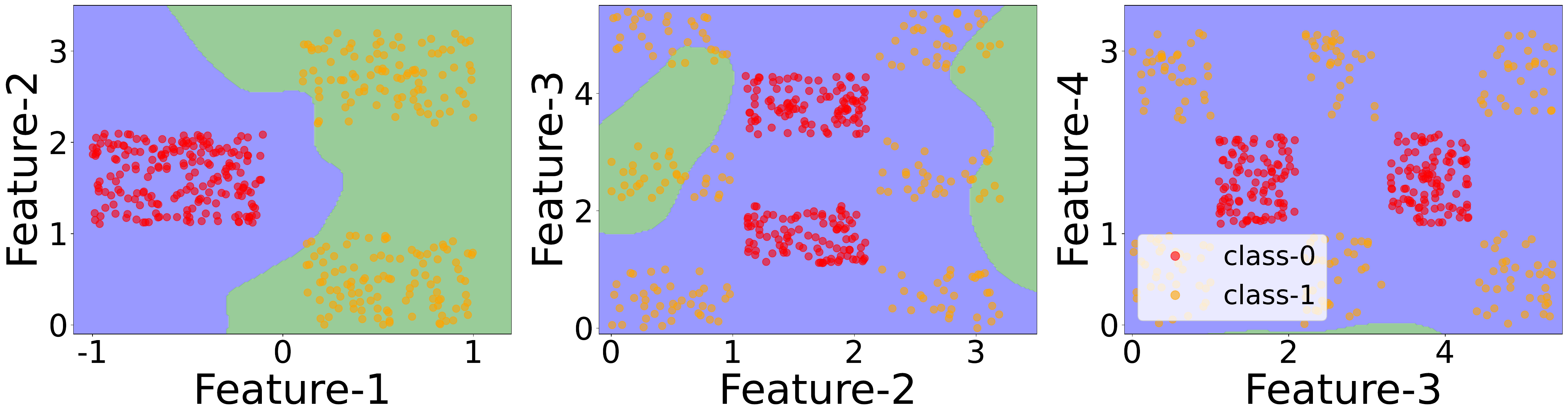}
    \end{subfigure}

\caption{\textbf{Diverse Decision boundary} plots for different random initializations, Each subfigure shows decision boundary across 4 dimension for pre-training with noisy random labels.}
        \label{fig15:show-all_random_init}
\end{figure}

\begin{figure}[t]
        \centering
        
         \begin{subfigure}{0.7\textwidth}
      \centering
      \includegraphics[width=\textwidth]{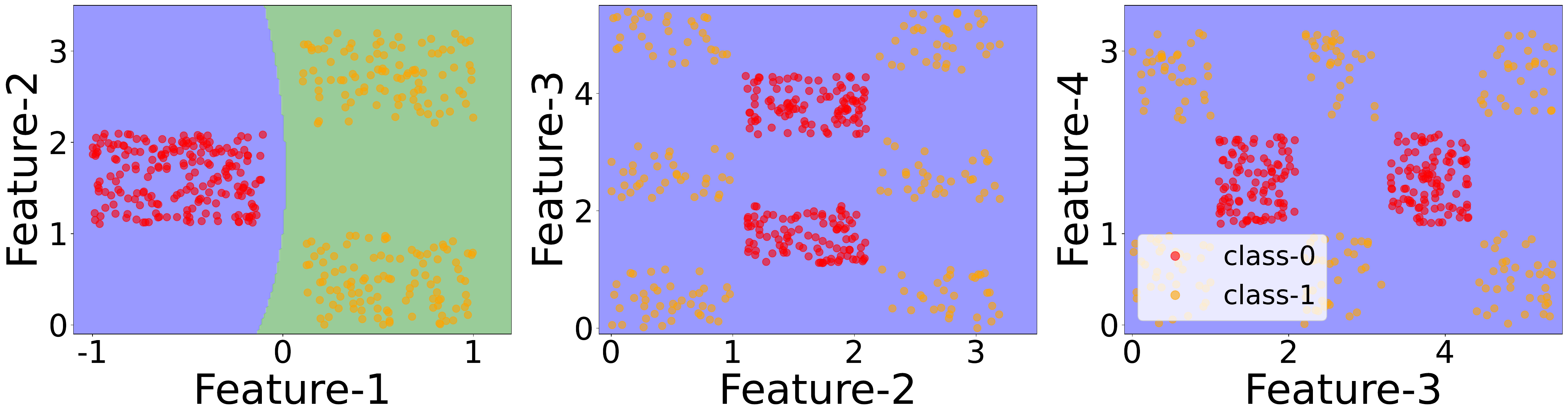}
    \end{subfigure}

     \begin{subfigure}{0.7\textwidth}
      \centering
      \includegraphics[width=\textwidth]{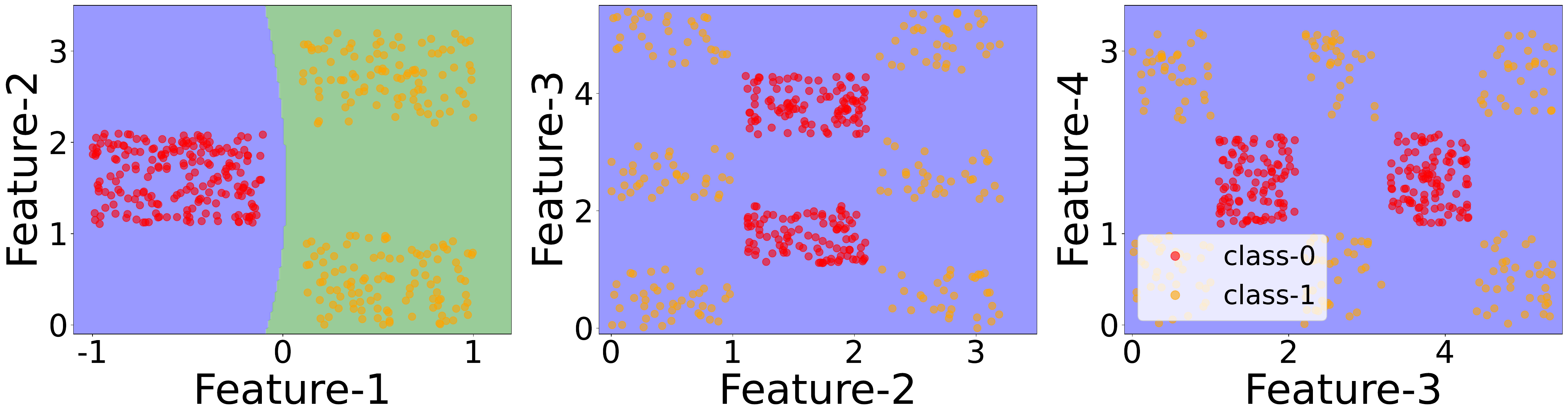}
    \end{subfigure}

    \begin{subfigure}{0.7\textwidth}
      \centering
      \includegraphics[width=\textwidth]{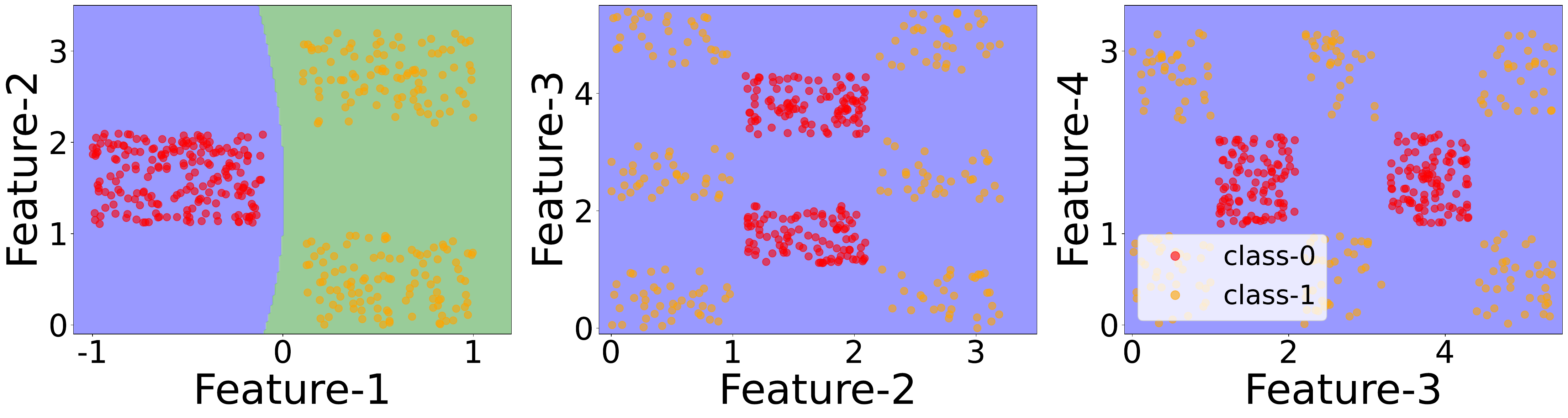}
    \end{subfigure} 
    
  \begin{subfigure}{0.7\textwidth}
      \centering
      \includegraphics[width=\textwidth]{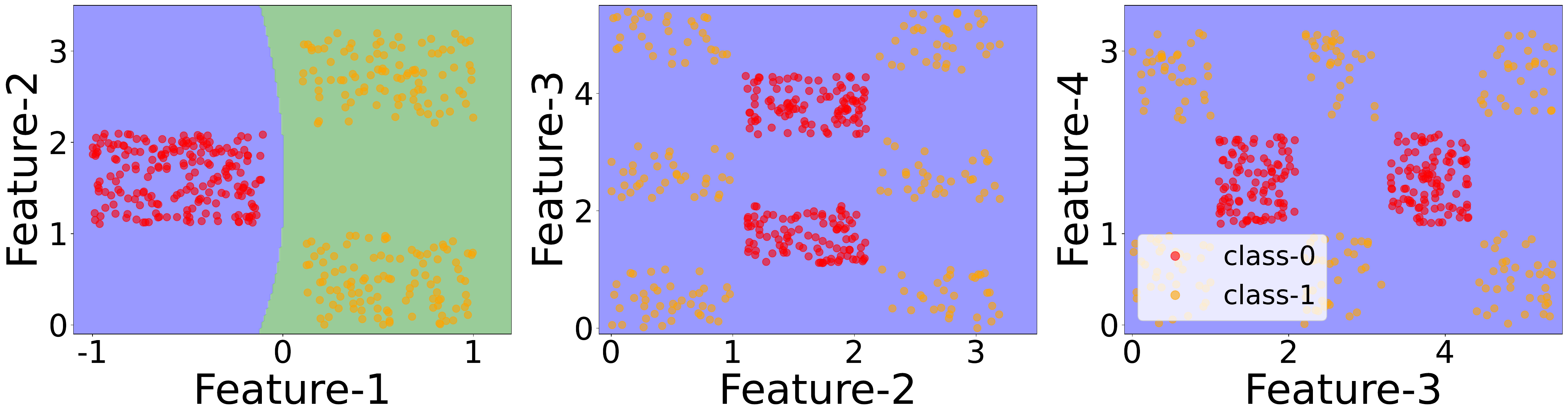}
    \end{subfigure}

    \begin{subfigure}{0.7\textwidth}
      \centering
      \includegraphics[width=\textwidth]{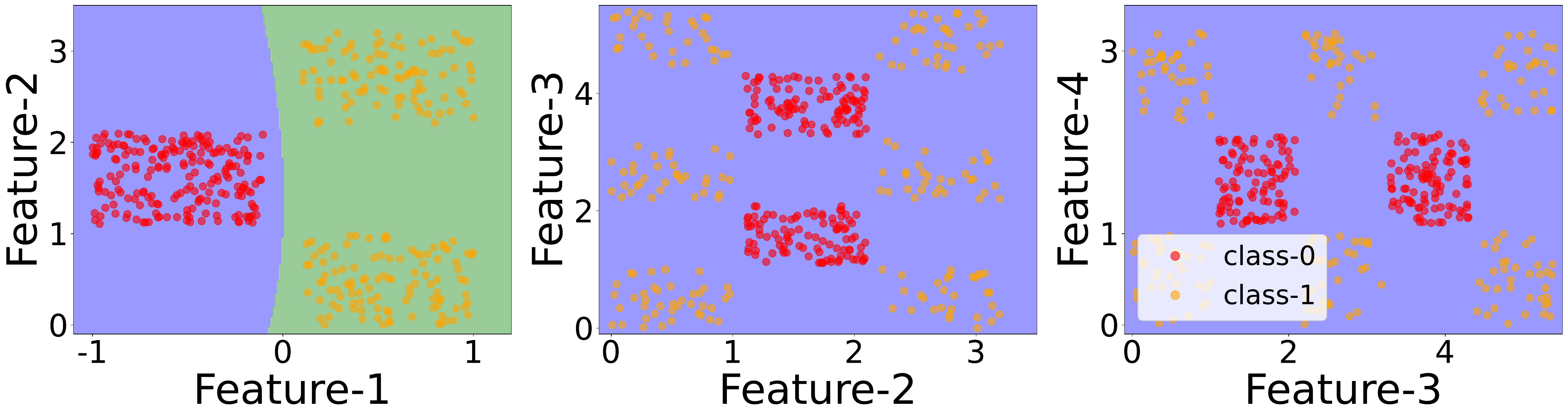}
    \end{subfigure}

  \begin{subfigure}{0.7\textwidth}
      \centering
      \includegraphics[width=\textwidth]{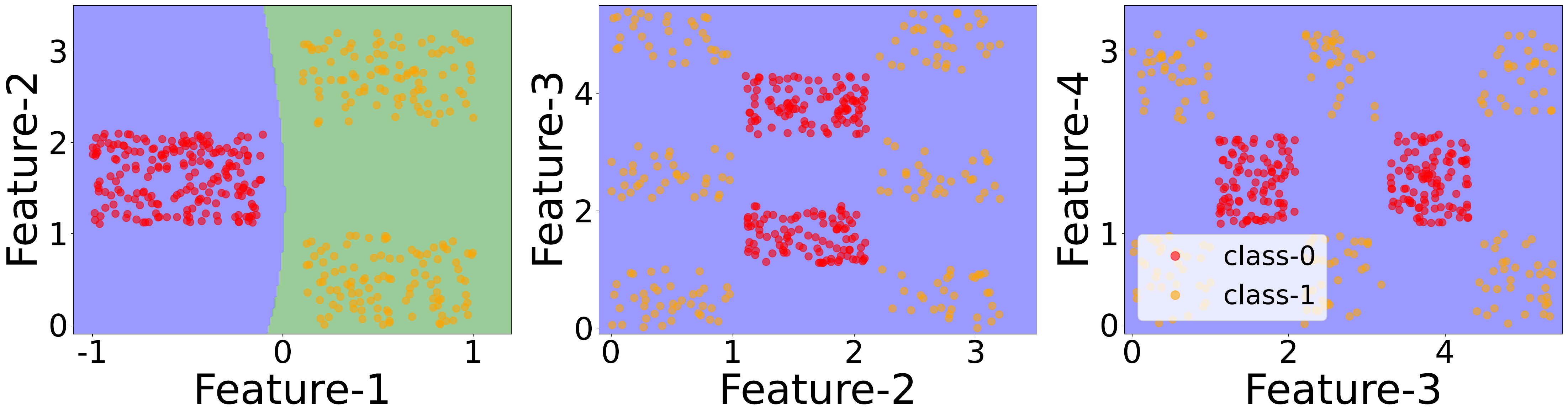}
    \end{subfigure}

       \caption{ \textbf{Similar Decision boundary } plots for different random initializations, Each subfigure shows decision boundary across 4 dimension for training with correct label (without any pretraining).}
       \label{fig16:show-all_random_init} 
\end{figure}

In this section, we show that under standard training, the model converges to a similar family of decision functions, which are easier to learn. This convergence does not happen under pretrained noise based training, which allows model to learn diverse set of functions with reliance on harder to learn features. Figure \ref{fig15:show-all_random_init} and \ref{fig16:show-all_random_init} demonstrates this finding using decision boundary plots for 4d slab data described in Section \ref{sec:method}. 
 
\clearpage
\subsection{Impact of Label Smoothing}
\label{sec:appdx-smoothing}
Label smoothing  \cite{DBLP:journals/corr/SzegedyVISW15} is a regularization technique, which replaces the one-hot ground truth vectors, with mixture of ground truth vectors and uniform distribution. In case of noisy pretraining, flip the labels of a fraction of data samples, whereas in case of label smoothing, the ground truth labels are mixture of one hot vector and uniform distribution, which act as a addition of noise to labels, thus showing equivalent effect.

\begin{table}[h]
\begin{tabular}{|c|c|ccc|ccc|}
\hline
\multirow{2}{*}{\begin{tabular}[c]{@{}c@{}}Data\\ \end{tabular}} &
  \multirow{2}{*}{\begin{tabular}[c]{@{}c@{}}LS \\ coeff.\end{tabular}} &
  \multicolumn{3}{c|}{Standard Training} &
  \multicolumn{3}{c|}{Noisy Pre-training} \\ \cline{3-8} 
 &
   &
  \multicolumn{1}{c|}{Train Acc.} &
  \multicolumn{1}{c|}{In-group} &
  Out-group &
  \multicolumn{1}{c|}{Train Acc.} &
  \multicolumn{1}{c|}{In-group} &
  Out-group \\ \hline
\multirow{2}{*}{$\data$} &
  0.0 &
  \multicolumn{1}{c|}{\nostd{100}{0.0}} &
  \multicolumn{1}{c|}{\nostd{85.2}{0.43}} &
  \nostd{38.5}{0.88} &
  \multicolumn{1}{c|}{\nostd{99.5}{0.34}} &
  \multicolumn{1}{c|}{\nostd{78.1}{1.02}} &
  \nostd{44.1}{1.60} \\ \cline{2-8} 
 &
  0.2 &
  \multicolumn{1}{c|}{\nostd{100}{0.0}} &
  \multicolumn{1}{c|}{\nostd{84.4}{0.42}} &
  \nostd{43.5}{3.62} &
  \multicolumn{1}{c|}{\nostd{99.8}{0.11}} &
  \multicolumn{1}{c|}{\nostd{77.3}{0.75}} &
  \nostd{51.1}{2.17} \\ \hline
\multirow{2}{*}{$\data'$} &
  0.0 &
  \multicolumn{1}{c|}{\nostd{100}{0.0}} &
  \multicolumn{1}{c|}{\nostd{84.1}{0.48}} &
  \nostd{44.4}{0.67} &
  \multicolumn{1}{c|}{\nostd{99.5}{0.73}} &
  \multicolumn{1}{c|}{\nostd{77.8}{1.15}} &
  \nostd{46.9}{0.92} \\ \cline{2-8} 
 &
  0.2 &
  \multicolumn{1}{c|}{\nostd{100}{0.0}} &
  \multicolumn{1}{c|}{\nostd{83.4}{0.49}} &
  \nostd{49.1}{3.63} &
  \multicolumn{1}{c|}{\nostd{99.5}{0.11}} &
  \multicolumn{1}{c|}{\nostd{78.5}{0.28}} &
  \nostd{52.2}{1.29} \\ \hline
\end{tabular}
 \vspace{0.1in}
\caption{\textbf{WaterBirds Dataset}: Test In-group and Out-group accuracies averaged over 10 runs on waterbirds dataset, with 100\% ($\mathcal{D}$) and 95\% ($\mathcal{D}'$) correlation of background with the true labels. For noisy pre-training, we corrupt $10\%$ of labels. 
Label smoothing (LS) varies between 0.0 and 0.2}  \label{tbl:appdx-waterbird}
\end{table}

\begin{table}[h]
\centering
\begin{tabular}{|l|lll|}
\hline
\multicolumn{1}{|c|}{\multirow{2}{*}{Parameter Initialization Regime}} & \multicolumn{3}{c|}{Feature Shuffle Accuracy}                                                                            \\ \cline{2-4} 
\multicolumn{1}{|c|}{}                       & \multicolumn{1}{c|}{Feature-1}           & \multicolumn{1}{c|}{Feature-2}          & \multicolumn{1}{c|}{Feature-3} \\ \hline
Label Smoothing ($0.2$) for feature confidence                             & \multicolumn{1}{l|}{\nostd{75.0}{4.4}}  & \multicolumn{1}{l|}{\nostd{92.9}{3.9}} & \nostd{91.5}{3.4}             \\ \hline
\end{tabular}
 \vspace{0.1in}
\caption{\textbf{Multi-Slab Data:} Randomized feature shuffle accuracies, averaged over $10$ runs. All models are trained to achieve $100$\% training and test accuracy. 
 } \label{tbl:appdx-multi-slab}
\vspace{-0.2in}
\end{table}

\subsection{Varying Model Architecture}
\begin{table}[h]
\centering
\begin{tabular}{|c|c|ccc|ccc|}
\hline
\multirow{2}{*}{\begin{tabular}[c]{@{}c@{}} Data\\ \end{tabular}} &
  \multirow{2}{*}{\begin{tabular}[c]{@{}c@{}}Model\\ Layer\end{tabular}} &
  \multicolumn{3}{c|}{Standard Training} &
  \multicolumn{3}{c|}{Noisy Pre-training} \\ \cline{3-8} 
 &
   &
  \multicolumn{1}{c|}{Train Acc.} &
  \multicolumn{1}{c|}{Top Rnd.} &
  Bottom Rnd. &
  \multicolumn{1}{c|}{Train Acc.} &
  \multicolumn{1}{c|}{Top Rnd.} &
  Bottom Rnd. \\ \hline
\multirow{2}{*}{$\data$} &
  3 &
  \multicolumn{1}{c|}{\nostd{99.9}{0.0}} &
  \multicolumn{1}{c|}{\nostd{52.5}{0.33}} &
  \nostd{98.3}{0.05} &
  \multicolumn{1}{c|}{\nostd{99.7}{0.07}} &
  \multicolumn{1}{c|}{\nostd{53.6}{1.56}} &
  \nostd{88.6}{0.76} \\ \cline{2-8} 
 &
  4 &
  \multicolumn{1}{c|}{\nostd{100}{0.0}} &
  \multicolumn{1}{c|}{\nostd{52.6}{0.24}} &
  \nostd{98.1}{0.05} &
  \multicolumn{1}{c|}{\nostd{99.8}{0.03}} &
  \multicolumn{1}{c|}{\nostd{55.6}{1.50}} &
  \nostd{87.6}{0.81} \\ \hline
\multirow{2}{*}{$\data'$} &
  3 &
  \multicolumn{1}{c|}{\nostd{98.5}{0.07}} &
  \multicolumn{1}{c|}{\nostd{93.1}{0.33}} &
  \nostd{56.5}{0.42} &
  \multicolumn{1}{c|}{\nostd{99.9}{0.06}} &
  \multicolumn{1}{c|}{\nostd{81.2}{1.02}} &
  \nostd{57.2}{1.50} \\ \cline{2-8} 
 &
  4 &
  \multicolumn{1}{c|}{\nostd{100}{0.0}} &
  \multicolumn{1}{c|}{\nostd{92.5}{0.17}} &
  \nostd{56.8}{0.08} &
  \multicolumn{1}{c|}{\nostd{100}{0.0}} &
  \multicolumn{1}{c|}{\nostd{84.4}{0.74}} &
  \nostd{56.9}{1.30} \\ \hline
\end{tabular}
 \vspace{0.1in}
    \caption{\textbf{Dominoes MNIST-FMNIST}: 
    Randomized shuffle accuracies, averaged over 3 runs, with 100\% ($\mathcal{D}$) and 95\% ($\mathcal{D}'$) correlation of MNIST with the true labels. MNIST:FMNIST. 
    Each for 3 and 4 layer fully connected networks. Noisy pre-training uses $10\%$ corrupt labels.  }  \label{tbl:appex-dominoes}
    	\vspace{-0.2in}
\end{table}

\clearpage
\subsection{Visualizing Feature Importance Using Top Eigenvectors}
\label{sec:appndx_eigen}

\begin{figure}[h]
    \centering
    \begin{subfigure}{0.32\linewidth}
        \includegraphics[width=\linewidth, height=8cm]{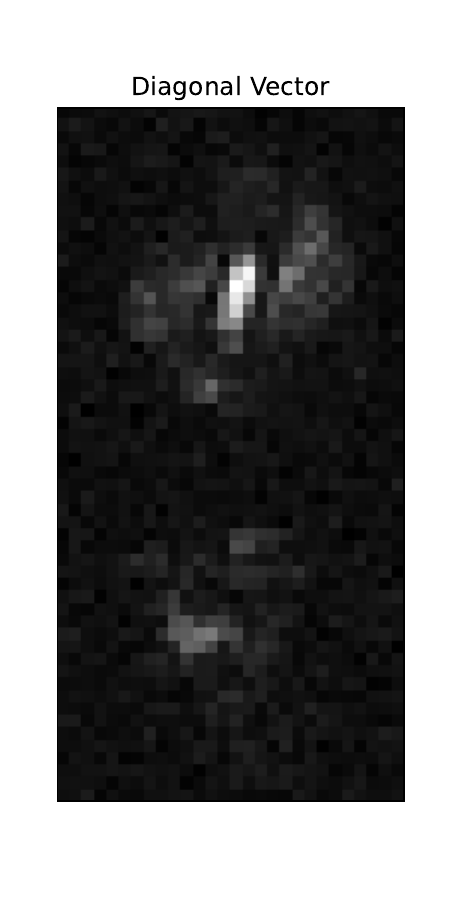}
        \caption{$\data$ standard training)}
        \label{fig:95-corr-00-noise}
    \end{subfigure}
    \hfill
    \begin{subfigure}{0.32\linewidth}
        \includegraphics[width=\linewidth, height=8cm]{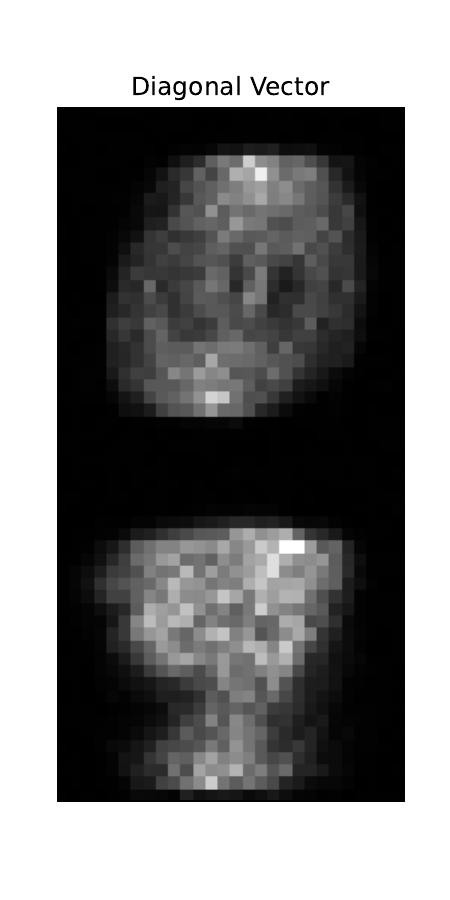}
        \caption{$\data'$ noisy pre-training}
        \label{fig:95-corr-100-noise}
    \end{subfigure}
    \hfill
    \begin{subfigure}{0.32\linewidth}
        \includegraphics[width=\linewidth, height=8cm]{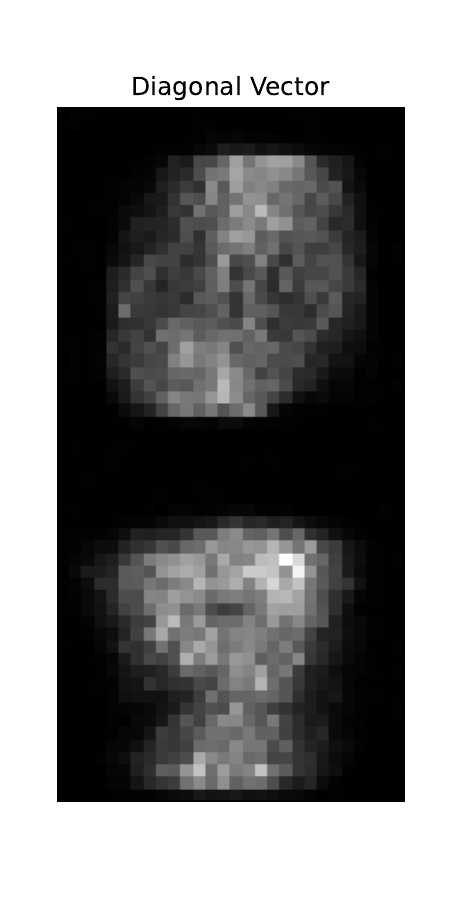}
        \caption{$\data$ noisy pre-training}
        \label{fig:100-corr-100-noise}
    \end{subfigure}
    \caption{Visualizations for diagonal of the first layer neural feature matrix $W_1^{\top}W_1$. 100\% ($\mathcal{D}$) and 95\% ($\mathcal{D}'$) correlation of instances with the true labels}
    \label{fig:first-three-images}
\end{figure}

\begin{figure}[h]
    \centering
    \begin{subfigure}{0.32\linewidth}
        \includegraphics[width=\linewidth, height=8cm]{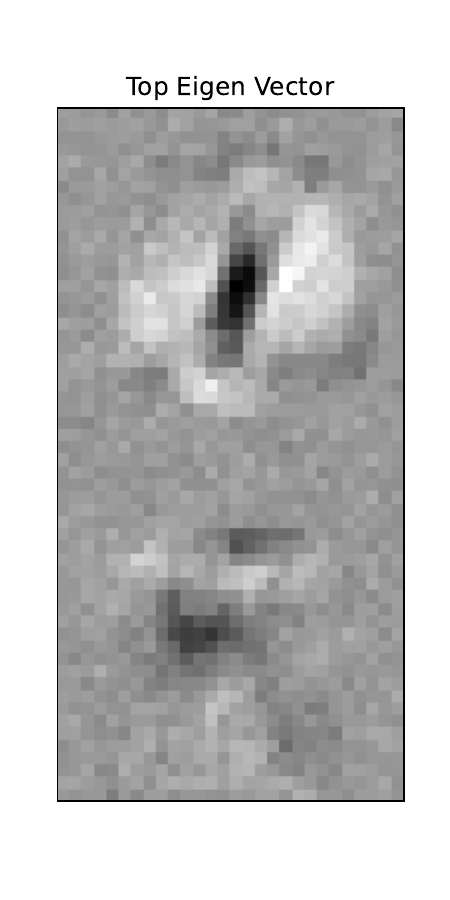}
        \caption{$\data$ standard training}
        \label{fig:100-corr-00-noise}
    \end{subfigure}
    \hfill
    \begin{subfigure}{0.32\linewidth}
        \includegraphics[width=\linewidth, height=8cm]{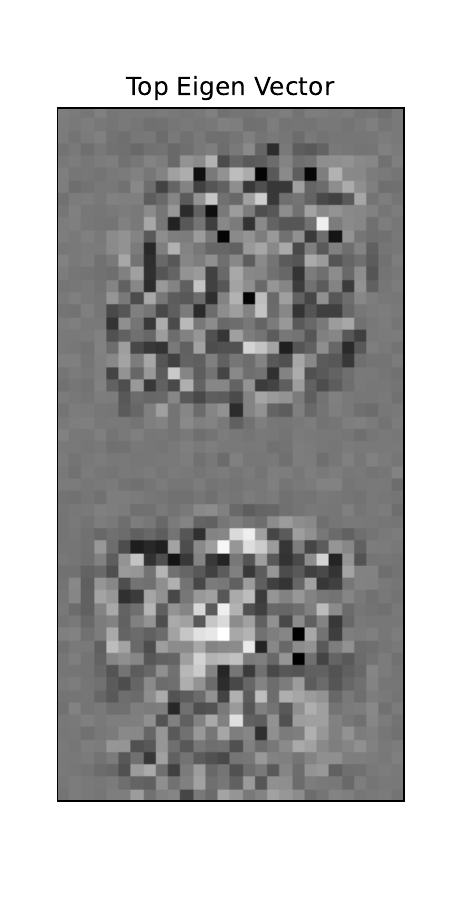}
        \caption{$\data'$ noisy pre-training}
        \label{fig:95-corr-100-noise}
    \end{subfigure}
    \hfill
    \begin{subfigure}{0.32\linewidth}
        \includegraphics[width=\linewidth, height=8cm]{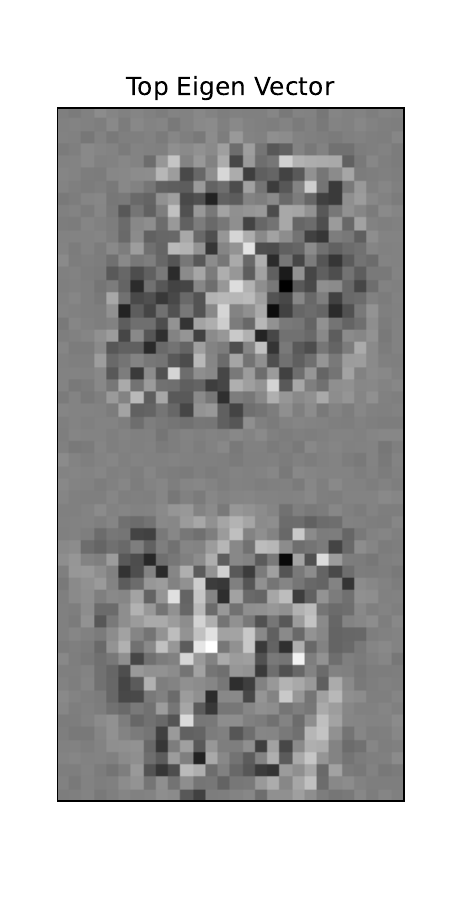}
        \caption{$\data$  noisy pre-training}
        \label{fig:100-corr-100-noise}
    \end{subfigure}
    \caption{Visualizations for top eigenvector of the first layer neural feature matrix $W_1^{\top}W_1$. 100\% ($\mathcal{D}$) and 95\% ($\mathcal{D}'$) correlation of instances with the true labels}
    \label{fig:second-three-images}
\end{figure}